\documentclass{article}

\usepackage[final,nonatbib]{neurips_2022}
\usepackage[utf8]{inputenc} 
\usepackage[T1]{fontenc}    
\usepackage{hyperref}       
\usepackage{url}            
\usepackage{booktabs}       
\usepackage{amsfonts}       
\usepackage{nicefrac}       
\usepackage{microtype}      
\usepackage{xcolor}         

\usepackage[ruled,vlined]{algorithm2e}
\usepackage{tabularx}
\usepackage{multirow}
\usepackage{makecell}
\usepackage{color, colortbl}
\usepackage[colorinlistoftodos, textwidth=20mm]{todonotes}
\usepackage{mathtools}

\def\x{{\boldsymbol x}}
\def\a{{\boldsymbol a}}
\def\w{{\boldsymbol w}}
\def\param{{\boldsymbol \theta}}
\def\y{{\boldsymbol y}}

\def\tco{{\it train-clean-100}}
\def\tct{{\it train-clean-360}}
\def\tof{{\it train-other-500}}
\def\tsmall{{\it train-10h}}
\def\devclean{{\it dev-clean}}
\def\devother{{\it dev-other}}
\def\testclean{{\it test-clean}}
\def\testother{{\it test-other}}

\definecolor{grey}{cmyk}{0.1,0.1,0.1,0.1}
\definecolor{orange}{cmyk}{0.1,0.7,0.5,0.2}

\newcommand{\librispeech}{{LibriSpeech}}
\newcommand{\librilight}{{Libri-Light}}

\title{Continuous Soft Pseudo-Labeling in ASR}

\author{
  Tatiana Likhomanenko, Ronan Collobert, Navdeep Jaitly, Samy Bengio \\
  Apple \\
  \texttt{\{antares,collobert,njaitly,bengio\}@apple.com} \\
}

\begin{document}

\maketitle
\begin{abstract}
Continuous pseudo-labeling (PL) algorithms such as slimIPL have recently emerged as a powerful strategy for semi-supervised learning in speech recognition. In contrast with earlier strategies that alternated between training a model and generating pseudo-labels (PLs) with it, here PLs are generated in end-to-end manner as training proceeds, improving training speed and the accuracy of the final model.  PL shares a common theme with teacher-student models such as distillation in that a teacher model generates targets that need to be mimicked by the student model being trained. However, interestingly, PL strategies in general use hard-labels, whereas distillation uses the distribution over labels as the target to mimic.  Inspired by distillation we expect that specifying the whole distribution (aka soft-labels) over sequences as the target for unlabeled data, instead of a single best pass pseudo-labeled transcript (hard-labels) should improve PL performance and convergence. 
Surprisingly and unexpectedly, we find that soft-labels targets can lead to training divergence, with the model collapsing to a degenerate token distribution per frame. 
We hypothesize that the reason this does not happen with hard-labels is that training loss on hard-labels imposes sequence-level consistency that keeps the model from collapsing to the degenerate solution. 
In this paper, we show several experiments that support this hypothesis, and experiment with several regularization approaches that can ameliorate the degenerate collapse when using soft-labels. These approaches can bring the accuracy of soft-labels closer to that of hard-labels, and while they are unable to outperform them yet, they serve as a useful framework for further improvements.

\end{abstract}

\section{Introduction and Related Works}
The Deep Learning community has recently made significant progress in semi-supervised learning via self-training or pseudo-labeling (PL)~\cite{scudder1965probability} in vision, speech and other domains~\cite{sohn2020fixmatch,berthelot2019mixmatch,berthelot2019remixmatch,chen2020semi,xu2020iterative,likhomanenko2020slimipl,manohar2021kaizen,higuchi2021momentum,higuchi2022advancing,mpl2022}. In this paper we focus on self-training in the context of automatic speech recognition (ASR). ASR presents some unique challenges in self-training that are not seen in some other tasks stemming from their sequential nature. Inputs and targets have different lengths and targets are not produced with the same regularity, compared to the inputs. 
Further, ASR models are often coupled with language models, that can be additionally used to aid the pseudo-labeling process. 

In semi-supervised learning~\cite{chapelle2006alexanderzien} we have access to labeled and unlabeled data, where the unlabeled data is typically much larger, since it is often easier to find unlabeled data, compared to labeled data. PL, in its simplest formulation, employs a base model trained with labeled data which is later used as a ``teacher'' to generate predictions on the unlabeled data. The unlabeled data is converted to ``pseudo-labeled'' data through the addition of these teacher generated targets. A student model is then trained on the combination of labeled and pseudo-labeled data to yield a final model.

PL with ASR has recently been applied to three different types of end-to-end models -- Connectionist Temporal Classification (CTC)~\cite{graves2006connectionist}, Transducer~\cite{graves2012sequence}, and sequence-to-sequence~\cite{chan2016listen}. 
To achieve state-of-the-art results, PLs are generated from the models using an external language model (LM) either with beam-search decoding~\cite{hsu2020semi,xu2020iterative,xu2020self} or through shallow fusion~\cite{kahn2020self,synnaeve2019end,park2020improved,zhang2020pushing}.
Unlike in vision, here PLs are sequences of tokens, such as characters\footnote{We refer to them as hard-labels because only some sequences are used and not the whole distribution over all sequences.}. With this setting, however, acoustic models tend to overfit to the text training set of the LM used for PL~\cite{xu2020iterative,park2020improved,mpl2022}.

Recent approaches on PL in ASR  have attempted to rectify this problem through continuous training~\cite{lee2013pseudo,chen2020semi}, in which the model that is being trained is the same model that is used to generate PLs, and PLs are generated without using an LM. The main challenge for continuous PL is stability, which was reported independently by several authors~\cite{likhomanenko2020slimipl,higuchi2021momentum,higuchi2022advancing,mpl2022} and resolved partially by introducing either a caching mechanism~\cite{likhomanenko2020slimipl} which kept an evolving cache of PL data or by using an Exponential Moving Average (EMA) of the model being trained to generate the PLs~\cite{manohar2021kaizen,higuchi2021momentum,higuchi2022advancing,mpl2022}.  This continuous training greatly simplifies the pipeline, improves convergence speed and achieves state-of-the-art compared to other semi-supervised approaches~\cite{likhomanenko2020slimipl}.

Interestingly, the above strategies in general use one (or a small number) of PLs from the teacher to train the student. By contrast distillation~\cite{hinton2015distilling} and soft-labeling \cite{berthelot2019mixmatch,berthelot2019remixmatch} approaches use the predicted distribution over labels from the teacher as the target to mimic. Inspired by these works we expect that specifying the whole distribution (aka soft-labels) over sequences as the target for unlabeled data should improve performance and convergence. 
However, with CTC-based models for labeled data we find experimentally that training with soft-labels leads to models where the distribution converges to a degenerate solution, e.g. emitting only the blank token (this only happens in continuous pseudo-labeling, while teacher-student exhibit stable soft-labeling, see Appendix~\ref{sec:app-teacher-student}).
By contrast, hard-labels are more stable and collapse to a degenerate solution is not observed. 
We believe this is because structured CTC loss used for hard-labels imposes sequence-level consistency while 
loss on soft-labels (cross-entropy or $L_2$) does not have this property, could destroy the sequence-level representation and thus easily converge to the degenerate solution, quite bad for generalization.
We apply different regularization approaches to try and prevent degenerate solutions such as entropy regularization, smoothing of the soft-labels, sampling from the teacher distribution~\cite{imamura2018enhancement,edunov2018understanding}, blending the soft-labels with hard-labels and PLs filtering. These approaches can bring the accuracy of soft-labels closer to that of hard-labels, and while they are unable to outperform them yet, they serve as a useful framework for further improvements.

\section{Background on Pseudo-Labeling and slimIPL}
\label{sec:pl-slimipl}

Let $L = \{\x_i, \y_i\}$ and $U = \{\x_j\}$ be the labeled and unlabeled datasets respectively. 
We consider a PL approach where an acoustic model $\mathcal{A}(\x ; \param)$ with model parameters $\param$ is continuously trained on a combination of $L$ and a pseudo-labelled set derived from $U$.
The model is trained by minimizing a loss $\mathcal{L}(\param) = \mathcal{L}_L(\param) + \lambda \mathcal{L}_U(\param)$ where $\lambda \in\mathbb{R}^+$ is a tunable hyper-parameter controlling the importance of unlabeled data. 
The loss for labeled data is defined as  $\mathcal{L}_L(\param) = - \displaystyle \mathbb{E}_{(\x,\y) \sim L} \log p_{\param}(\y|\x)$,  where $p_\param(\y|\x)$ is the conditional distribution defined by model $\mathcal{A}(\x;\param)$.
The loss for unlabeled data is defined as $\mathcal{L}_U(\param) = - \displaystyle \mathbb{E}_{\x \sim U}  \log  p_{\param}(\hat{\y} |\x)$, where $\hat{\y}$ is the PL transcription for a data point generated using the model being trained and a prior LM. Specifically,
\begin{equation}
\label{eq:hardlm}
\hat \y = \underset{\y}{\operatorname{argmax}} ~ \log p_{\param}(\y|\x) + \alpha \, \log p^{lm}(\y),\,\x\in U,
\end{equation}
where $p^{lm}(\y)$ is an LM, and $\alpha \ge 0$ is a hyper-parameter controlling the reliance on the LM.
The above minimization is usually approximated with a beam-search decoding algorithm~\cite{synnaeve2019end}. 
In recent works it was found that $\alpha > 0$ gives stable training but can lead to over-fitting to the language model used~\cite{xu2020iterative} while $\alpha=0$ can lead to training instability and divergence~\cite{likhomanenko2020slimipl}. 

To resolve these issues an LM free ($\alpha=0$) continuous PL approach called slimIPL was recently proposed for ASR~\cite{likhomanenko2020slimipl}. 
It also optimizes further the speed of the PL generation: instead of Eq.~\eqref{eq:hardlm} with $\alpha=0$ which defines hard-transcriptions, slimIPL uses hard-alignment defined in Eq.~\eqref{eq:hard-a}. Models trained with the CTC loss output distributions over token set $\w$ (including an auxiliary blank token) for every output frame, or timestamp, $t=\overline{1,T}$ as $p_\param^t(w|\x), w\in\w$, where $T$ is the total number of frames in the sample. 
The CTC loss maximizes the probability $p_{\param}(\y|\x)$ of a transcription $\y$ across all possible alignments\footnote{E.g. for $T=12$ one of  the alignments for $\y$ `cat' is `cc\#\#\#aatttt\#', where \# is a CTC blank token.} $\a=(a_1,\dots,a_T)$ which lead to $\y$ assuming factorization $p_{\param}(\y|\x)=\sum_{\a\to\y} \prod_{t=1}^T p_\param^t(a_t|\x)$.
In order to get PLs generated by Eq.~\eqref{eq:hardlm} ($\alpha=0$) we would need to compute the total probability of the transcription $\y$ by summing probabilities of every alignment $\a\to\y$ and then taking the most probable $\y$. 
This is intractable but beam search could be used in this case in order to approximate the argmax search. 
With slimIPL, the authors decided to use a simpler approach~-- hard-alignment (or hard-path), obtained with:
\begin{equation}
    \label{eq:hard-a}
    \hat{\y} = \mathcal{G}(\hat{\a}), \qquad \hat{\a}=(\hat{a}_1,\dots, \hat{a}_T), \qquad  \hat{a}_i=\operatorname{argmax}_{w}{p_\param^i(w|\x)}\,,
\end{equation}
where for every frame the most probable token is $\hat{a}_i$. From there, the final pseudo-label transcription $\hat{\y}$ is obtained through the transformation denoted by $\mathcal{G}$, which removes repeated tokens as well as the CTC blank token.\footnote{This can be viewed as hard-alignment, obtained from per-frame hard-labels. In contrast, CTC decoding considers all alignments leading to the same transcription to obtain this hard-alignment.} This is equivalent to minimizing Eq. (\ref{eq:hardlm}) with a (lexicon-free) beam search constrained to a beam size of 1.

A schematic overview of slimIPL is presented in Appendix~\ref{app:slimipl} Algorithm~\ref{algo:softslimipl}. slimIPL exploits a dynamic cache, which cleverly facilitates access to PLs from previous model states (this can be viewed as model ensemble for PL generation).
This dynamic cache was proposed in order to stabilize the optimization and to avoid sudden model divergence due to collapsing to empty transcription predictions. 
slimIPL has several hyper-parameters: (i) $\lambda$, (ii) $M$, the training step at which PL generation begins, (iii) $C$, the dynamic cache size and (iv) $p$, the probability of updating the cache.

Our goal in the next sections is to investigate how ``soft''-labeling can further improve slimIPL as we believe it may contain richer information compared to hard-labels.
Interestingly, ``soft''-labeling is computationally cheap, similar to the cost implied by hard-paths obtained from Eq.~\eqref{eq:hard-a}, and much cheaper than with pseudo-labels which would be obtained by minimizing Eq.~\eqref{eq:hardlm} (via a beam search) but requires more memory.
We start our analysis from the best models found in slimIPL work~\cite{likhomanenko2020slimipl}.

\section{Experimental Setup and Methods}\label{sec:soft}
Inspired by results on soft-labeling in vision~\cite{berthelot2019mixmatch,berthelot2019remixmatch} and knowledge distillation~\cite{hinton2015distilling} our empirical hypothesis is the following: exploiting the whole distribution over sequences should be more powerful for PL and bring benefit in either faster convergence and / or lower error at the end of training. In this section we describe the data, the training setup and methods used to test this hypothesis.

{\bf Data}~~~
We use the \librispeech{} dataset~\cite{panayotov2015librispeech} and the \tsmall{} subset of \librilight{}~\cite{librilight}.
We use the \tct{} and \tof{} standard \librispeech{} subsets as unlabeled data and consider either 10h (\tsmall{}) or 100h (\tco{}) as the labeled data.
The standard sets \devclean{} and \devother{} are used to tune all hyper-parameters and to select the best models. The sets \testclean{} and \testother{} are used to report final token (TER) and word (WER) error rates.
\vspace{0.2cm}\\
{\bf Training setup}~~~We closely follow slimIPL experiments~\cite{likhomanenko2020slimipl} with the same hyper-parameters, including transformer model, CTC loss and character tokens (see Appendix~\ref{app:setting}).
We make only few changes: a) to speed up training and decrease memory usage we get rid of relative positional embedding~\cite{shaw2018self} and use CAPE~\cite{likhomanenko2021cape} instead; b) we train with larger batch per GPU but on 8 GPUs.
In our experiments we only vary $C$, $p$ and the type of PL (soft-labels vs hard-alignment), while keeping everything else fixed. 
We run experiments with 3 seeds and we report mean and standard deviation as the final performance.
\vspace{0.2cm}\\
{\bf Soft-labeling slimIPL}~~~
Even though speech recognition involves a sequence-to-sequence alignment problem, we are going to perform ``soft''-labeling on a per-frame basis. In that respect, the soft-label strategy is similar than what is commonly used for non-sequential data.
Formally, let's denote logits as $z^t_w=\log p^t_\param(w|\x)$ and PL logits $\hat{z}^t_w=\log p^t_{\param'}(w|\x)$ for $\x \in U$. We also denote as $q^t_w(\x; \tau) = \frac{e^{z_w^t/\tau}}{\sum_{w'} e^{z_{w'}^t/\tau}}$ and $\hat{q}^t_w(\x; \tau)=\frac{e^{\hat z_w^t/\tau}}{\sum_{w'} e^{\hat z_{w'}^t/\tau}}$. Then the loss function on unlabeled data $\x \in U$ is defined in a similar way than the distillation loss~\cite{hinton2015distilling}: either as cross-entropy (CE) with temperature $\tau > 0$ or as a $L_2$ regression (equivalent to CE when $\tau=\infty$, see~\cite{hinton2015distilling}):
\begin{equation}
  \small
  \label{eq:soft-entropy-loss}
    \mathcal{L}_{U, ce}(\param) = - \beta\displaystyle \mathbb{E}_{\x \sim U} \sum_{t,w} q^t_w \log{\hat{q}^t_w}; \qquad
    \mathcal{L}_{U, L_2}(\param) = - \beta\displaystyle \mathbb{E}_{\x \sim U} \sum_{t,w} (z^t_w - \hat z^t_w)^2.
\end{equation}  
In order to keep the same $\lambda$ (ratio between labeled and unlabeled data losses, introduced in Section~\ref{sec:pl-slimipl}) for both hard-labels and soft-labels experiments, we add a re-scaling factor $\beta > 0$ in Eq.~\eqref{eq:soft-entropy-loss}, which provides a way to keep gradients computed for $\mathcal{L}_U$ and $\mathcal{L}_{U, ce}$/$\mathcal{L}_{U, L_2}$ at the same scale.\footnote{As we use adaptive optimizer, we tried also to use two different optimizers for different losses as their dynamics are different. However, experiments with two optimizers were significantly worse, not to mention it brings complexity in practice.} 

\section{Empirical Study and Discussion}

\begin{table*}[t!]
\caption{
Comparison between different PL generation ways with their best hyper-parameters: ``hard-path'' denotes Eq.~\eqref{eq:hard-a} PLs, ``hard-beam'' denotes approximation of Eq.~\eqref{eq:hardlm} ($\alpha=0)$ via beam search, ``soft'' denotes PLs from Sec.~\ref{sec:soft}. Upper and lower bounds denote training on labeled data only.
\label{tab:resuts}}
\begin{footnotesize}
\begin{center}
\setlength\tabcolsep{4pt} 
\resizebox{0.95\linewidth}{!}{
\begin{tabular}{@{}lccccccccc@{}}
\toprule
\multirow{2}{*}{PL type} & Cache & \multicolumn{2}{c}{PL vs golden label} & \multicolumn{2}{c}{PL vs hard-alignment} &  \multicolumn{4}{c}{WER \%} \\
\cmidrule(lr){3-4} \cmidrule(lr){5-6} \cmidrule(lr){7-10}
& $C$ & TER \% & WER \% & TER \% & WER \% & dev-clean & dev-other & test-clean & test-other \\
\midrule
 lower bound (sup. LS) & - & -  & - & - & - & 2.6 $\pm$ 0.1 & 6.9 $\pm$ 0.1 & 2.7 $\pm$ 0.1 & 6.9 $\pm$ 0.1 \\
 \midrule
 100h, hard-path & 1k & 1.5 $\pm$ 0.1 & 4.8 $\pm$ 0.1 & 0.0 & 0.0 & 3.9 $\pm$ 0.1 & 8.6 $\pm$ 0.1 & 4.1 $\pm$ 0.1 & 9.2 $\pm$ 0.2 \\
 100h, hard-path & 100 & 1.3 $\pm$ 0.1 & 4.2 $\pm$ 0.1 & 0.0 & 0.0 & 3.7 $\pm$ 0.1 & 8.0 $\pm$ 0.1 & 3.9 $\pm$ 0.1 & 8.2 $\pm$ 0.1 \\
 100h, hard-beam (beam 10) & 100 & 1.3 $\pm$ 0.1 & 4.1 $\pm$ 0.1 & 0.007 $\pm$ 0.001 & 0.04 $\pm$ 0.01 & 3.7 $\pm$  0.1 & 7.9 $\pm$  0.1 & 3.9 $\pm$  0.1 & 8.3 $\pm$ 0.1 \\
\midrule
100h, sampling ($\tau=0.4$) & 100 & 1.4 $\pm$ 0.1 & 4.2 $\pm$ 0.1 & 0.04 $\pm$ 0.01 & 0.2 $\pm$ 0.1 & 3.7 $\pm$ 0.1 & 8.0 $\pm$ 0.2 & 3.9 $\pm$ 0.1 & 8.3 $\pm$ 0.2 \\ 
\midrule
 100h, soft ($\tau=10,\beta=1$) & 1k & 1.6 $\pm$ 0.1 & 5.3 $\pm$ 0.3 & - & - & 4.7 $\pm$ 0.4  & 9.0 $\pm$ 0.5  & 4.8 $\pm$ 0.4  & 9.5 $\pm$ 0.5  \\
 \,\,\, +hard ($\delta=0.1$) & 100 & 1.4 $\pm$ 0.1 & 4.5 $\pm$ 0.1 & -  & - & 4.1 $\pm$ 0.1 & 8.6 $\pm$ 0.2 & 4.2 $\pm$ 0.1 & 9.2 $\pm$ 0.1 \\ 
 \,\,\, +hard ($\tau=1,\beta=0.01,\delta=0.1$) & 100 & 1.4 $\pm$ 0.1 & 4.4 $\pm$ 0.1 & - & - & 4.0 $\pm$ 0.1 & 8.3 $\pm$ 0.1 & 4.2 $\pm$ 0.1 & 8.7 $\pm$ 0.2 \\ 
 \midrule
 \midrule
 10h, hard-path & 1k & 4.3 $\pm$ 0.1 & 14.7 $\pm$ 0.3 & - & - & 14.4 $\pm$ 0.3 & 18.8 $\pm$ 0.4 & 15.1 $\pm$ 0.4 & 19.3 $\pm$ 0.3 \\
 10h, hard-beam (beam 100) & 1k & 3.8 $\pm$ 0.1 & 13.1 $\pm$ 0.1 & 0.008 $\pm$ 0.001 & 0.05 $\pm$ 0.01 & 13.0 $\pm$ 0.1 & 16.8 $\pm$ 0.2 & 13.5 $\pm$ 0.1 & 17.3 $\pm$ 0.1 \\
 \midrule
 10h, sampling ($\tau=0.15$) & 1k & 4.3 $\pm$ 0.3 & 14.7 $\pm$ 1.0 & 0.01 $\pm$ 0.02 & 0.08 $\pm$ 0.02 & 14.2 $\pm$ 0.8 & 18.5 $\pm$ 1.2 & 15.1 $\pm$ 0.9 & 19.1 $\pm$ 1.1 \\ 
 \midrule
 10h, soft ($\tau=1,\beta=0.01$) & 1k-3k & \multicolumn{8}{c}{we are not able to train, blowing up} \\
 \,\,\, +hard ($\delta=0.1$) & 1k & 4.6 $\pm$ 0.2 & 15.5 $\pm$ 0.2 & - & - & 14.9 $\pm$ 0.4 & 19.8 $\pm$ 0.2 & 15.7 $\pm$ 0.3 & 20.4 $\pm$ 0.5  \\
 \midrule
 upper bound (sup. 100h) & - & -  & - & - & - & 5.9 $\pm$ 0.1 & 17.9 $\pm$ 0.1 & 6.2 $\pm$ 0.1 & 18.1 $\pm$ 0.1 \\
 upper bound (sup. 10h) & - & -  & - & - & - & 37.1 $\pm$ 0.1 & 58.4 $\pm$ 0.1 & 37.7 $\pm$ 0.3 & 58.4 $\pm$ 0.2 \\
 \bottomrule
\end{tabular}
}
\end{center}
\end{footnotesize}
\end{table*}

\begin{figure}[t!]
    \centering
    \includegraphics[width=0.9\textwidth]{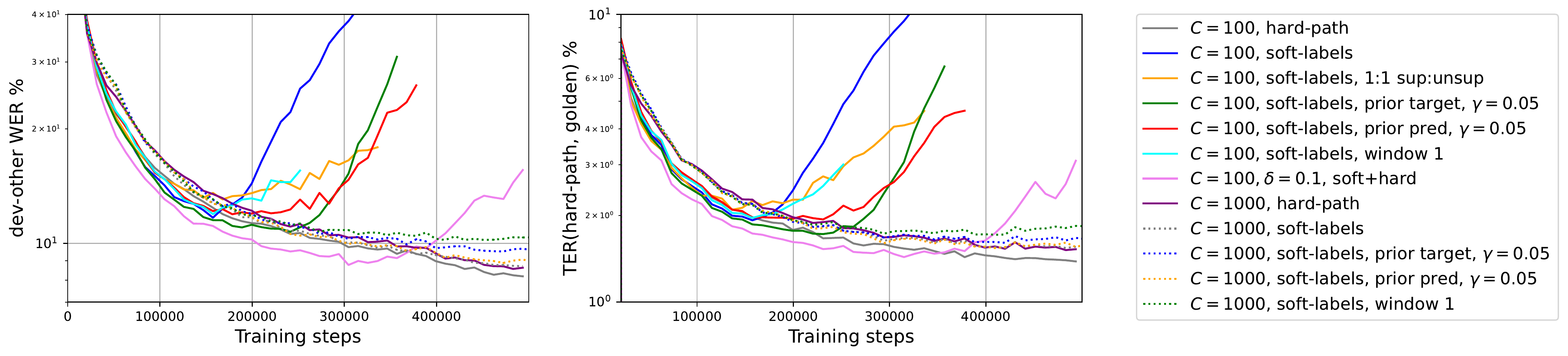}
    \caption{Comparison between different soft-labels for 100h supervision where additional regularizations are used, default parameters are $p=0.1, \tau=10, \beta=1, N_U=3$.}
    \label{fig:instability}
\end{figure}

\begin{figure}[t!]
    \centering
    \includegraphics[width=\textwidth]{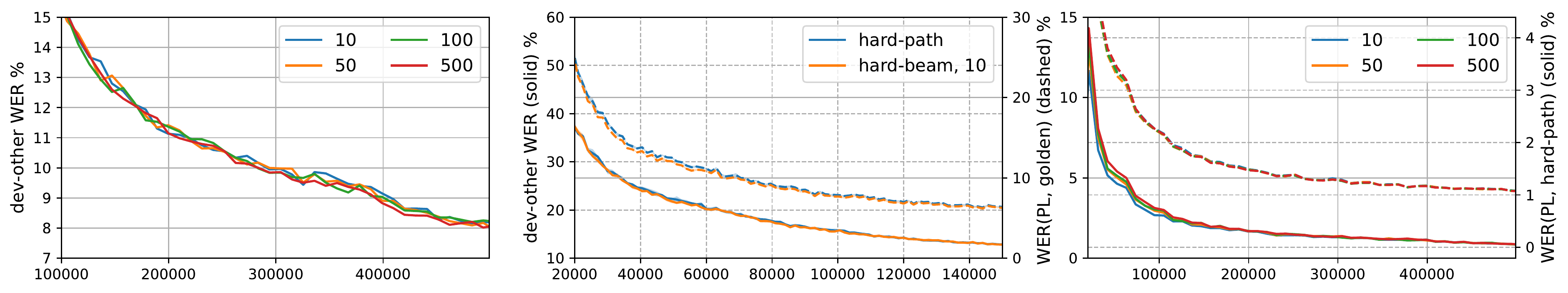}
    \includegraphics[width=\textwidth]{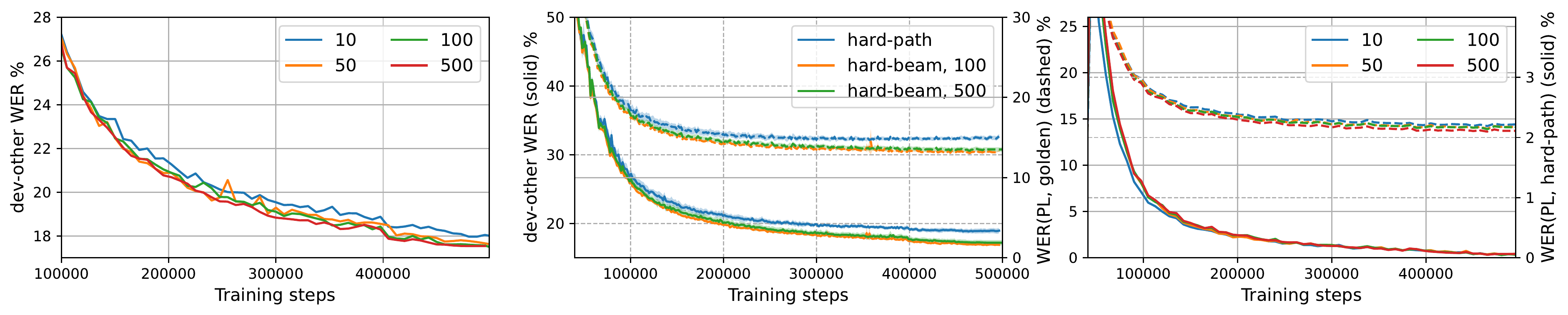}
    \caption{Comparison between hard-transcription approximated with a beam (hard-beam) of different size and hard-path for 100h (top) and 10h (bottom) of supervision. Left: \devother{} WER for different beams; middle: best hard-beam vs hard-path across 3 different seeds on \devother{} (solid) and their PL quality (dashed). Right: PL quality (dashed) and WER(PL, hard-path) (solid) for different beams.}
    \label{fig:beam-training}
\end{figure}

{\bf Training with soft-labels is unstable}~~~First experiments with both soft-labels losses reveal some training instability, when using the same cache size $C$ and probability $p$ than with hard-labels: training diverges, with WER increasing on validation sets as well. The WER between PLs and their corresponding golden transcription also diverges.
By increasing the cache size by 10x (from 100 to 1k) and lowering its probability (0.1) the training is however stabilized (Figure~\ref{fig:instability}). 
We performed similar experiments with EMA~\cite{manohar2021kaizen,mpl2022} and observed the same trend: the EMA decay factor needs to be increased (from 0.999 to 0.9999/0.99995) compared to the one used with hard-alignments. 
Previously it was reported that EMA and slimIPL are robust to hyper-parameters and that using a small cache or a larger value of $p$ would still lead to stable training, and similar WER results.
All our experiments demonstrate that this is not the case with soft-labeling.
Moreover, the best soft-labeling performing model ($\tau=10$ and $\beta=1$) has worse WER performance than when using hard-alignment, as shown in Table~\ref{tab:resuts}.\footnote{Temperatures $\tau$ and scaling factor $\beta$ were searched in the range $[0.0001,100]$. Different cache size were tried (10, 100, 1k, 10k). Cache probability was picked in (0.1, 0.5, 1).}
The same 1k cache for hard-alignment gives worse WER performance than with 100 cache (as PLs are then more slowly updated), but it is still better than with soft-labels.
\vspace{0.2cm}\\
{\bf Soft-labeling easily collapses to blank}~~~To understand instability issues in soft-labeling, we track how the PLs of a subset of the unlabeled samples evolve over the training with $C=100, p=0.1$, see Appendix~\ref{app:plsamples} Figure~\ref{fig:per-frame-3} (top). Interestingly, only several (2-3) tokens dominate every frame during the training. 
When the model diverges, e.g. the blank token probability increases over all frames, ultimately taking over the whole distribution mass of soft-labels. \vspace{0.2cm}\\
{\bf Hard-labels frame-level transitions are noisy. Sequence-level ones are smooth.}~~~
Compared to soft-labels, hard-path dynamics (Appendix~\ref{app:plsamples} Figure~\ref{fig:per-frame-1}) exhibit non-smooth transition between alignments. Instead of looking at each sample individually, we summarize the evolution of the same subset of unlabeled samples in Figure~\ref{fig:pl-change-hard} (for hard-labels) and Figure~\ref{fig:pl-change-soft} (for soft-labels). We track the average distance between the PLs generated for the same sample, for two consecutive iterations. We report the frame-level Levenshtein distance of these consecutive PLs, after converting them into hard-alignments $\hat{\a}$ (whether the PLs were soft or hard). We also report the Levenshtein distance at the sequence level $\hat{\y}$. When tracking individual samples, we observe more transition noise in hard-labels, than for soft-labels.
However, at the sequence level, hard-labels PLs transitions are significantly smoother as the training goes.
\vspace{0.2cm}\\
{\bf Sequence-level PLs may lead to better WER}~~~
We now consider Eq.~\eqref{eq:hardlm} with beam sizes from 10 to 500 to understand if a richer sequence-level representation can be helpful, by turning off the LM ($\alpha=0$).  
Figure~\ref{fig:beam-training} top shows that beam and hard-path perform similarly. Moreover, the WER between PLs generated from beam search and PLs generated from hard-path becomes smaller than 2\% (right plot) very quickly (after 30k updates). 
However, Figure~\ref{fig:beam-training} bottom, shows that if we switch to using only 10h of supervision, we observe a 1.5-2\% absolute decrease in WER from increasing the beam size to 50-100, while the PLs from beam and hard-path are still similar (WER is 1-3\%). 

Based on the above observations we decided to investigate several ways to improve soft-labeling: regularizing the soft-label distribution in the hope to avoid blank collapses, sampling as a way to regularize soft-labels transitions, and finally combining soft and hard-labels.

\subsection{CTC blank regularization does not solve instability issue}

To resolve the issue of collapsing to blank tokens we start with different regularization schemes to constrain the predictive distribution to be close to a specified prior distribution. 
All variants we tried slow down divergence, but were unable to improve on our best slimIPL results (Figure~\ref{fig:instability}). 
The different prior distribution we tried are as follows:\\
{\bf Entropy regularization:} This prior attempts to maximize the entropy of the predictive distribution. \\
{\bf Sequence-level prior}: The average predicted probability vector across all time steps for a sample is regularized to be close to (using KL-divergence) either (i) a uniform prior across tokens, or (ii) a token distribution computed from \tco{} text. Since text does not have a blank token, the prior for the blank token is estimated by a supervised model trained on \tco{}.\\
{\bf Frame-level prior}: Each frame is regularized to either match a target blank probability, or the predictions and/or targets as label-smoothed by averaging with a uniform distribution with weight $\gamma$, which is
varied from 0.01 to 0.1.
\\{\bf Logits averaging}: Logits from neighboring frames are smoothed with window $w=1 \cdots 10$ frames.

\subsection{Sampling as stable soft-labeling}
\begin{figure}[t!]
    \centering
    \includegraphics[width=\textwidth]{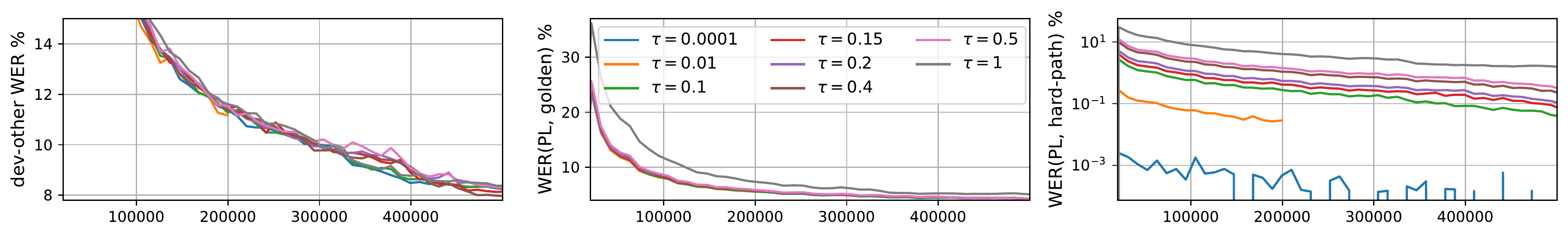}
    \includegraphics[width=\textwidth]{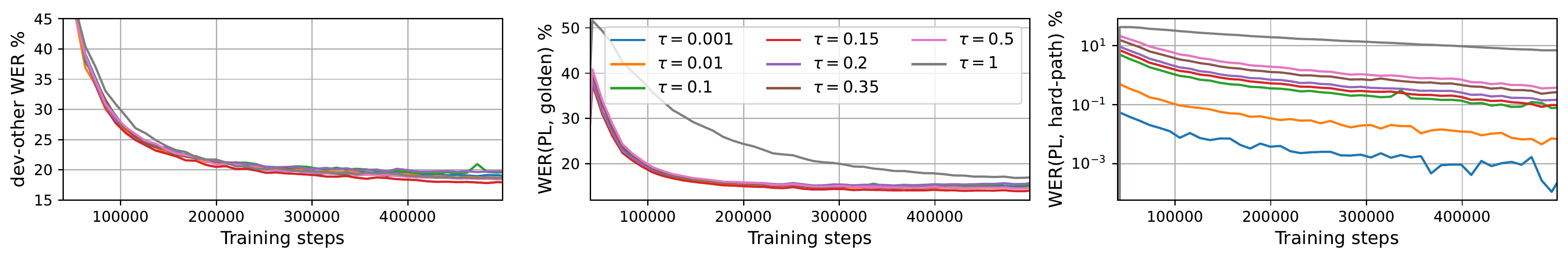}
    \caption{Sampling for 100h (top) and 10h (bottom) of supervision. WER on \devother{} (left) depending on temperature $\tau$ and WER between transcriptions obtained with sampling (and different temperature $\tau$) and golden transcription (left) / hard-alignment (middle).}
    \label{fig:sampling}
\end{figure}

Next, we investigate how sampling~\cite{imamura2018enhancement,edunov2018understanding} from the model performs because sampling encapsulates the uncertainty in the predictions. We generate $\hat{\y}=\mathcal{G}(\hat{a}_1, \dots, \hat{a}_T)$ where $\hat{a}_t \sim p^t_\param(\w|\x)$ for $t=\overline{1,T}$. A temperature $\tau$ is varied from 0.0001 to 10 to change the peakiness of the distribution, and CTC loss is used on PL data. 
For $\tau\to 0$ sampling corresponds to the hard-path while $\tau\to\infty$ corresponds to the uniform distribution over tokens.\footnote{With $\tau>1$ a distribution is closer to uniform and models do not train.}
We can see from Figure~\ref{fig:sampling} that sampling converges to hard-path in the end (top) while in limited supervision (bottom) only $\tau=1$ does not. As with hard-transcription we observe that sampling is similar to hard-path for 100h and is better for 10h of supervision. Unlike soft-labels, sampled PLs are stable like the hard-path PLs.

\subsection{Blending soft-labels with hard-labels as regularization}
We observed and discussed above, when models are trained with hard-alignments, hard-transcriptions or sampling, their frame-level predictions vary a lot from step to step during training. In contrast soft-labels do not
display such rapid changes in the predictive distribution (compare Appendix~\ref{app:plsamples} Figure~\ref{fig:pl-change-hard} vs Figure~\ref{fig:pl-change-soft}).
We hypothesize that soft-labeling loss is trivial to optimize to degenerate solutions and the model over-fits to it easily while memorizing the entire labeled set (see Appendix~\ref{sec:app-overfitting}).
Meanwhile, hard PL versions provide sequence level targets which might prevent a collapse. So we combine hard-path PL and soft-PL using $\delta \mathcal{L}_U + (1-\delta)\mathcal{L}_{U,ce}$, $\delta=\{0.01, 0.1, 0.5, 0.9, 0.99\}$. This is the most effective way to stabilize training: e.g. see Appendix~\ref{sec:app-blending} Figure~\ref{fig:blending-stability} with $\tau=1, C=100$.
Further, with this blend models can be trained with 10h of labeled data which was not possible with soft-labels alone (see Table~\ref{tab:resuts}).
The strategy also works better than the other soft-PL variants (Table~\ref{tab:resuts}) while converging 1.5-2x faster (see Figure~\ref{fig:instability}). 
However some issues still remain as it slowly starts to diverge at the very end of training, and the results are slightly worse than hard-path.

\section{Conclusion}
At first glance, a change from hard-labels to soft-labels for continuous pseudo-labeling in ASR seemed simple and obvious, but our experiments revealed entirely unexpected results.
Soft-labeling was found to quickly drive the PLs that are generated continuously to trivial, degenerate solutions while memorizing the entire labeled data at the same time.
We found that different regularization approaches help to improve stability of soft-label training but are unable to match the accuracy of hard-label trained models. 
Hard-transcription and sampling from the distribution on the other hand show stability and improved results in lower supervision setting.
We found that combining both hard-alignment PLs with CTC loss and soft-label PLs with cross-entropy loss on unsupervised data leads to improved results. 
This almost matches hard-alignment and speeds up convergence by 1.5-2x.

\bibliographystyle{IEEEtran}

\newpage
\appendix

\section{slimIPL Algorithm Overview}\label{app:slimipl}


\setlength{\textfloatsep}{5pt}
\begin{algorithm}[ht!]
\caption{slimIPL algorithm where arbitrary algorithm for PL generation is used}
\label{algo:softslimipl}
\SetAlgoLined
\KwData{labeled $L = \{\x_i, \y_i\}$ and unlabeled $U = \{\x_j\}$ data, $\tilde{\x} = augmentation(\x)$, initialization $\param^0$ and cache $\mathcal{C}=\{\}$, learning rate $\eta_k$, losses $\mathcal{L}_L$ and $\mathcal{L}_U$.}
\KwResult{Acoustic model $\mathcal{A}(\x; \param)$ }
    1. Train $\mathcal{A}$ on $(\x, \y)\in L$ for $M$ steps: $\param^{k+1} = \param^{k} - \eta_k \nabla \mathcal{L}_L(\mathcal{A}(\tilde{\x};\param^{k}), \y), k=\overline{1, M}$\;
    2. Decrease model's $\mathcal{A}(\x,\param)$ dropout\;
    3. \For{$k=\overline{M+1, M + C}$}{
        For random $\x\in U$ generate $\hat\y=PL(\mathcal{A}_{inference}(\x;\param^k))$ and $\mathcal{C} \leftarrow \mathcal{C} \bigcup \{\x, \hat \y\}$ \;
        $\param^{k+1} = \param^{k} - \eta_k \nabla \mathcal{L}_L(\mathcal{A}(\tilde{\x};\param^{k}), \y), (\x, \y)\in L$\;
    }
  \Repeat{convergence or maximum iterations are reached}{
    \eIf{$rand(0, 1) <N_L/(N_L + N_U)$}
        {Take random $(\x, \y)\in L$ and $\param^{k+1} = \param^{k} - \eta_k \nabla \mathcal{L}_L(\mathcal{A}(\tilde{\x};\param^{k}), \y)$}
        {
            Take random $b=(\x, \y)\in \mathcal{C}$ and $\param^{k+1} = \param^{k} - \eta_k \nabla \mathcal{L}_U(\mathcal{A}(\tilde{\x};\param^{k}), \y)$\;
            \If{$rand(0, 1)<p$}{
            For random $\x'\in U$ generate $\hat\y'=PL(\mathcal{A}_{inference}(\x;\param^k))$ and $\mathcal{C} \leftarrow \mathcal{C}\setminus b  \bigcup \{\x', \hat \y'\} $ 
            }
        }
        $k \leftarrow k+1$\;
  }
\end{algorithm}
\begin{table*}[h]
\caption{
Comparison between slimIPL and fully supervised prior works: the same model architecture (transformer with learnable relative positional embedding), loss and token set is used. This gives lower and upper bounds on the WER one can expect with semi-supervised approaches.
\label{tab:resuts-prior}}
\begin{footnotesize}
\begin{center}
\setlength\tabcolsep{5pt} 
\resizebox{0.75\linewidth}{!}{
\begin{tabular}{@{}lcrrrr@{}}
\toprule
\multirow{2}{*}{} & \multirow{2}{*}{Cache $C$} & \multicolumn{4}{c}{WER \%} \\
\cmidrule{3-4} \cmidrule{5-6}
& & dev-clean & dev-other & test-clean & test-other \\
\midrule
 lower bound (sup. 960h)~\cite{likhomanenko2020rethinking} & - & 2.6 & 7.0 & 2.7 & 6.8 \\
 \midrule
 slimIPL 100h, hard-path~\cite{likhomanenko2020slimipl} & 100 & 3.7 & 7.3 & 3.8 & 7.5 \\
 upper bound (sup. 100h)~\cite{likhomanenko2020slimipl} & - & 6.2 & 16.8 & 6.2 & 16.8 \\
 \midrule
 slimIPL 10h, hard-path~\cite{likhomanenko2020slimipl} & 1000 & 11.4 & 14.0 & 11.4 & 14.7 \\
 upper bound (sup. 10h)~\cite{likhomanenko2020slimipl} & - & 31.9 & 52.3 & 32.6 & 52.4 \\
 \bottomrule
\end{tabular}
}
\end{center}
\end{footnotesize}
\end{table*}

\section{Empirical Setup}\label{app:setting}

We keep the original 16kHz sampling rate from \librispeech{}, and compute log-mel filterbanks with 80 coefficients for a 25ms sliding window, strided by 10ms. 
All features are normalized to have zero mean and unit variance per input sequence before feeding them into the acoustic model.
We closely follow the architecture and training pipeline from slimIPL, including same hyper-parameters, CTC loss, transformer-based architecture, token set (26 English alphabet letters augmented with the apostrophe and a word boundary token), SpecAugment~\cite{park2019specaug} parameters, Adagrad optimizer~\cite{duchi2011adaptive} and dynamic batching. 
We make only few changes: a) to speed up the training and decrease memory usage, we replace relative positional embedding~\cite{shaw2018self} by CAPE positional embedding~\cite{likhomanenko2021cape} (only a global shift of 30s is used); b) we train all models on 8 A100 40GB GPUs with dynamic batching of $\sim 290$ audio per GPU, and FP32 tensor core computations for up to 500k updates. 
In all experiments we only vary the PL type and few slimIPL hyper-parameters (cache size $C$ and cache probability update $p$), while keeping everything else fixed.
Most of experiments are run with 3 different seeds, and we report mean and standard deviation as the final performance. We start with models where $C=100, M=20,000, N_U=3, p=0.1$ for 100h and $C=1000, M=60,000, N_U=10, p=0.1$ for 10h of supervision.

\newpage

\section{Frame-Level and Sequence-Level Evolution in PLs}\label{app:plsamples}

\begin{figure}[ht!]
    \centering
    \includegraphics[width=\textwidth]{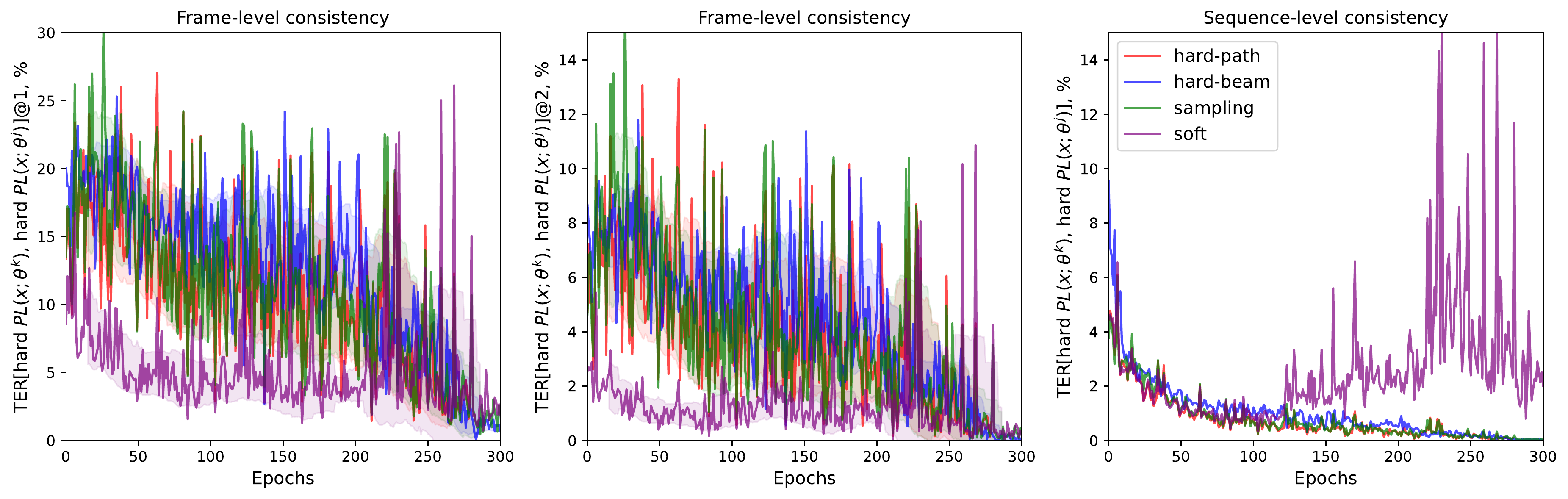}
    \caption{Levenshtein distance at the frame level (first and second columns) and sequence level (third column) between PLs for the same subset of unlabeled data for two consecutive steps for hard-path hard-beam, sampling and soft PLs.}
    \label{fig:pl-change-hard}
\end{figure}

\begin{figure}[ht!]
    \centering
    \includegraphics[width=\textwidth]{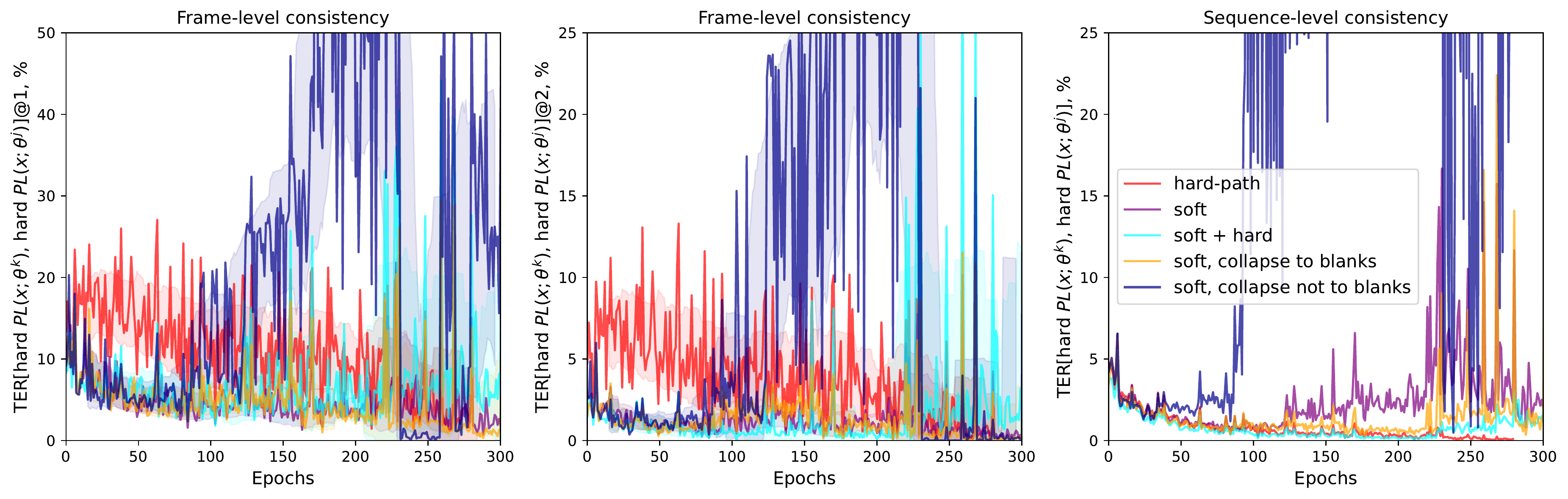}
    \caption{Levenshtein distance at the frame level (first and second columns) and sequence level (third column) between PLs for the same subset of unlabeled data for two consecutive steps for hard-path, soft,  soft blended with hard (``soft + hard'') soft blended with hard diverged to blank only (``soft, collapse to blanks''), soft with prior on target diverged to uniform distribution (``soft, collapse not to blanks'') PLs.}
    \label{fig:pl-change-soft}
\end{figure}

To further analyse convergence to a degenerate solution for soft-labels,
we keep tracking of PLs for the same subset of unlabeled samples as training goes (200 samples).
We then compute the average distance between the PLs generated for the same sample, for two consecutive iterations.
We report the frame-level Levenshtein distance of these consecutive PLs, after converting them into hard-alignments $\hat{\a}$ (whether the PLs were soft or hard). 
Here we consider also top-2 distance: we check whether $\hat{\a}_t^{j}$ is in top-2 tokens set of $p^t_{\param^k}(w|\x)$, where $k$ and $j$ are two consecutive iterations, and $k < j$.
We also report the Levenshtein distance at the sequence level $\hat{\y}$.
Results (for cache $C=100$ and $p=0.1$) are shown for hard-path, beam and sampling in Figure~\ref{fig:pl-change-hard}, and for different variants of soft-labels with regularization in Figure~\ref{fig:pl-change-soft}. 
When tracking individual samples, we observe more transition noise in hard-labels, than for soft-labels.
However, at the sequence level, hard-labels PLs transitions are significantly smoother as the training goes.
Moreover, for models converged to a degenerate solution (Figure~\ref{fig:pl-change-soft}) we observe first sequence-level and later also frame-level consistency destruction between PLs for two consecutive iterations.

\begin{figure}[ht!]
    \centering
    \includegraphics[width=\textwidth]{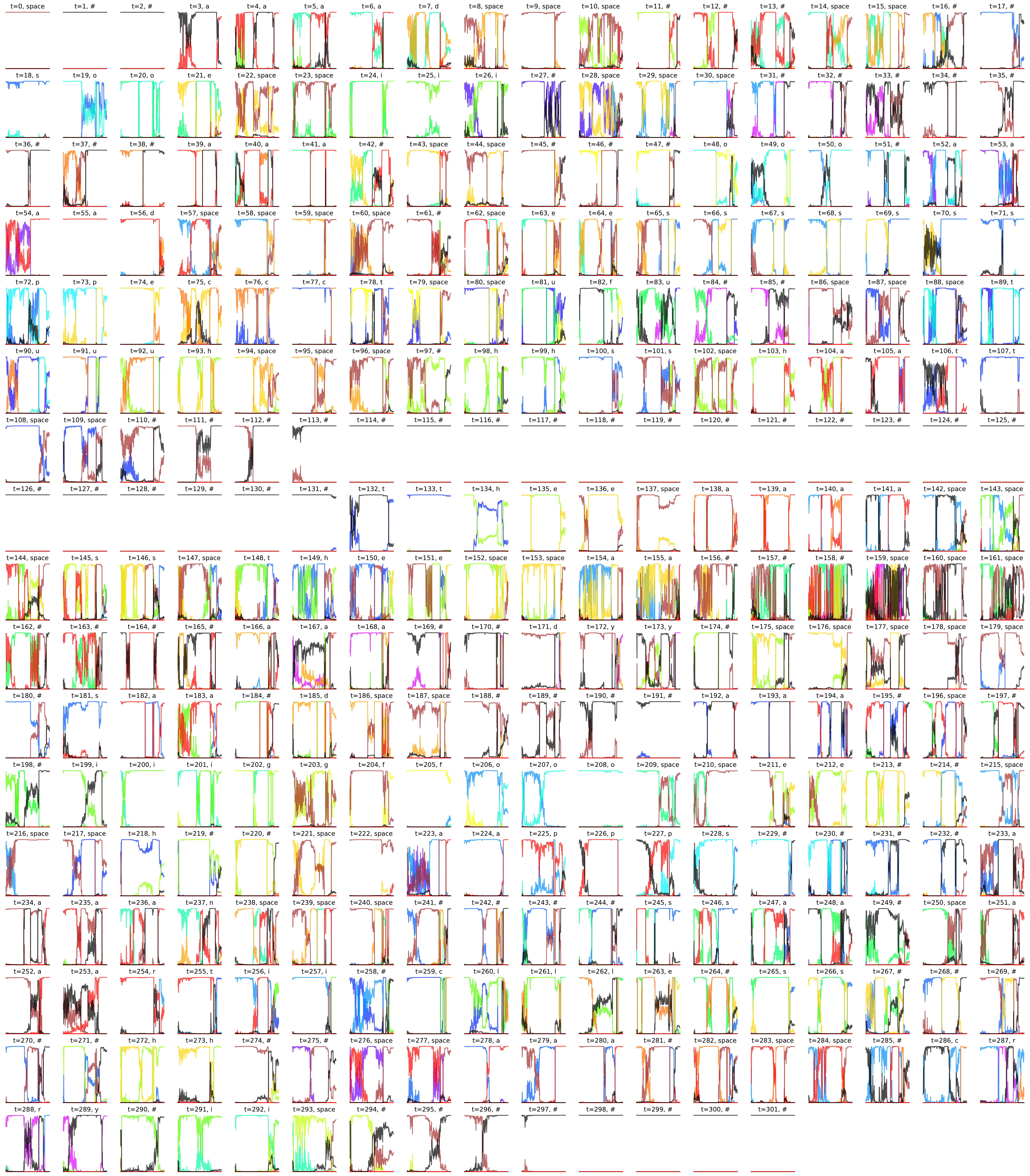}
    \includegraphics[width=\textwidth]{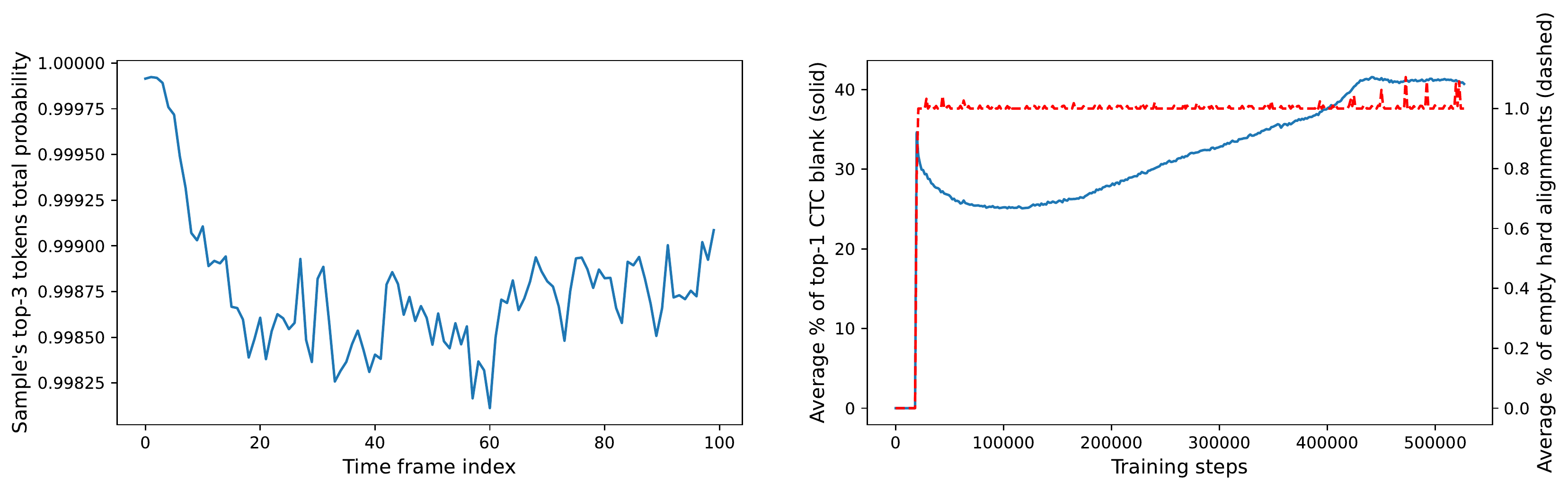}
    \caption{Training dynamics for soft-labels with 100h labeled data, $C=100, p=0.1, \tau=10, \beta=1$.}
    \label{fig:per-frame-3}
\end{figure}

\begin{figure}[ht!]
    \centering
    \includegraphics[width=\textwidth]{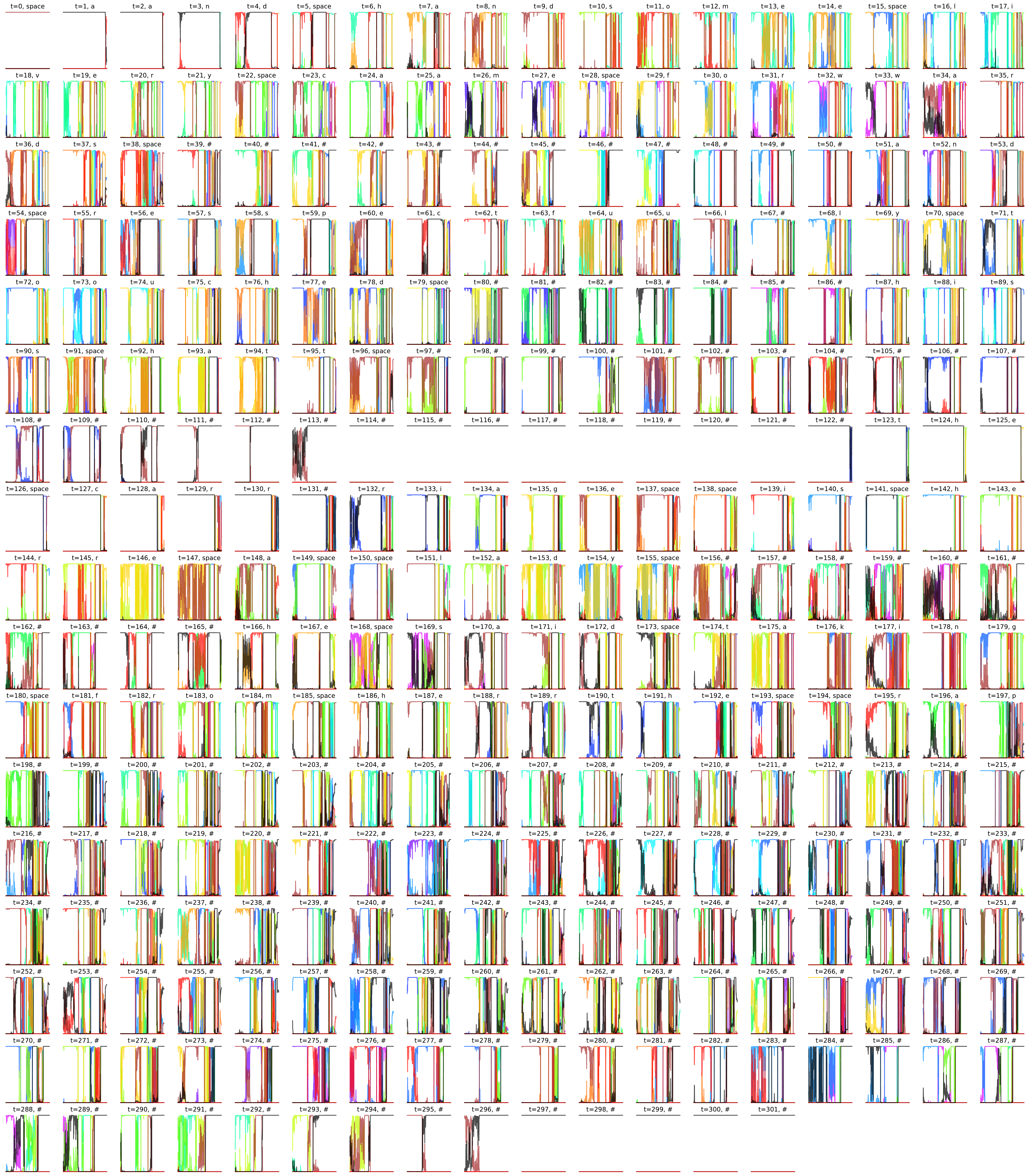}
    \includegraphics[width=\textwidth]{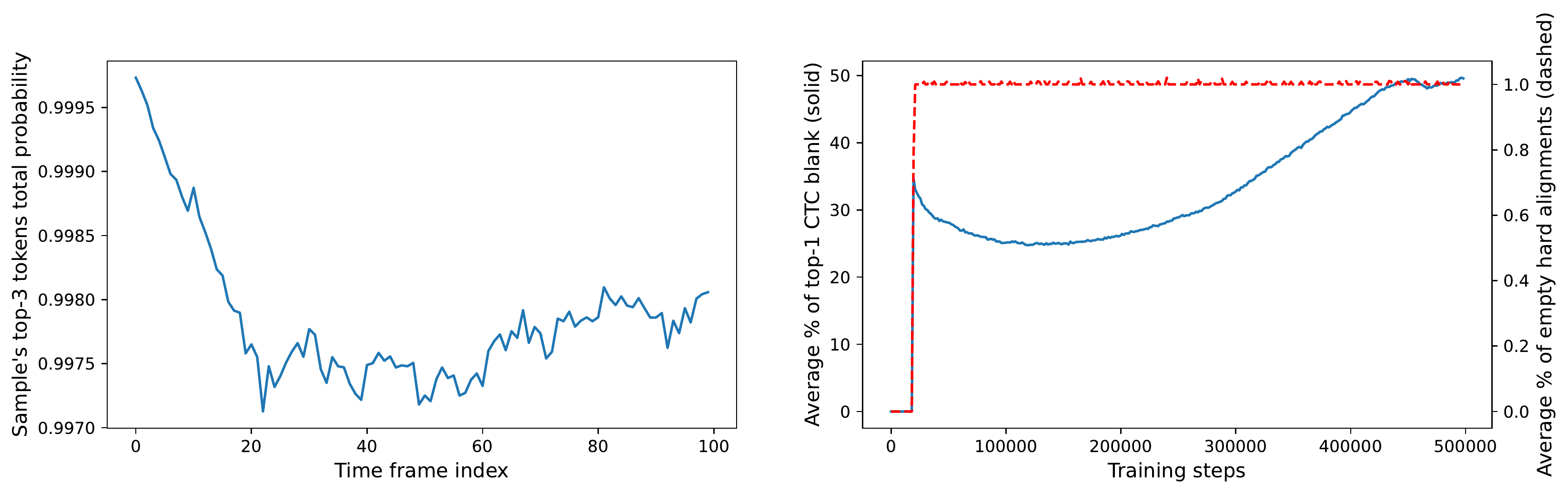}
    \caption{Training dynamics for best model with blending hard and soft-labels. We use 100h labeled data, $C=100, p=0.1, \tau=10, \beta=1, \delta=0.1$.}
    \label{fig:per-frame-4}
\end{figure}

\begin{figure}[ht!]
    \centering
    \includegraphics[width=\textwidth]{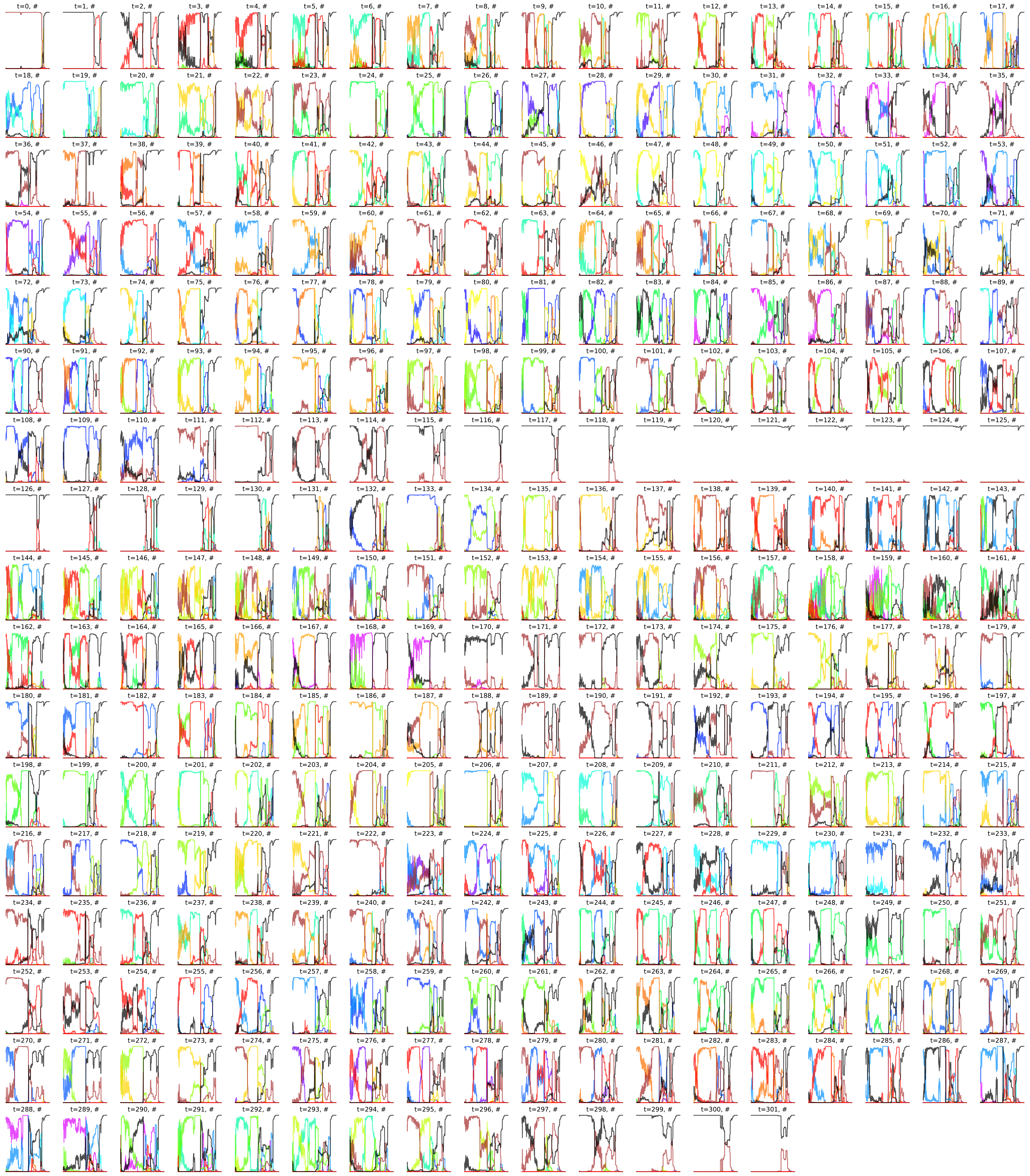}
    \includegraphics[width=\textwidth]{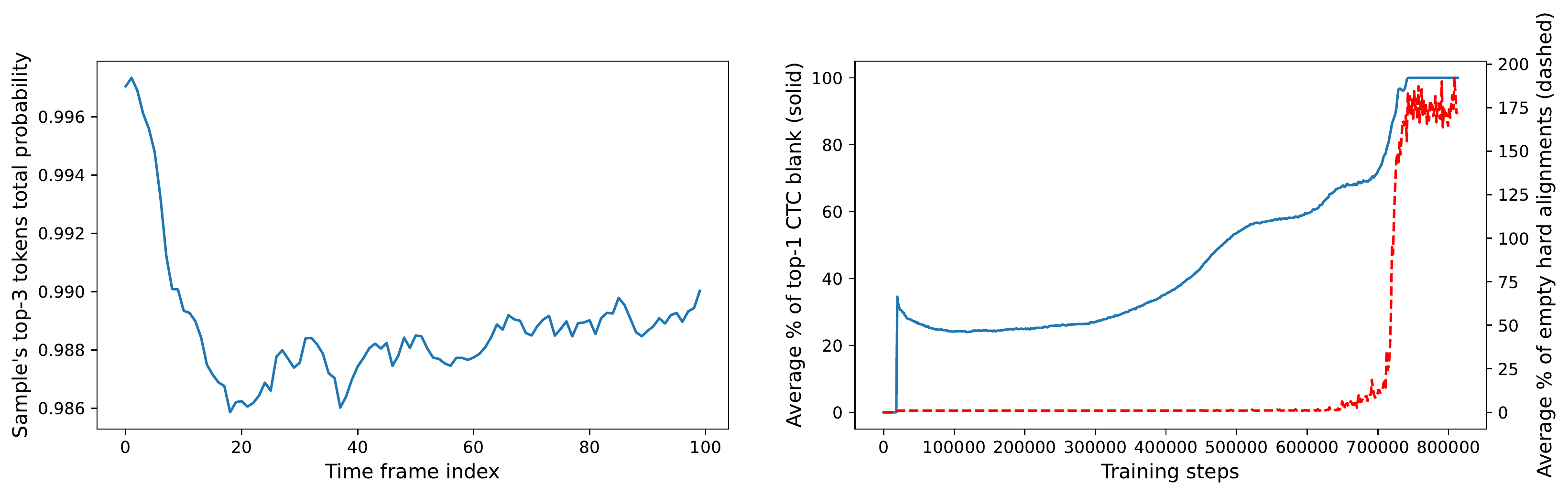}
    \caption{Training dynamics for blending hard and soft-labels together when divergence to blank distribution happens. We use 100h labeled data, $C=100, p=0.1, \tau=10, \beta=1, \delta=0.1$.}
    \label{fig:per-frame-5}
\end{figure}

\begin{figure}[ht!]
    \centering
    \includegraphics[width=\textwidth]{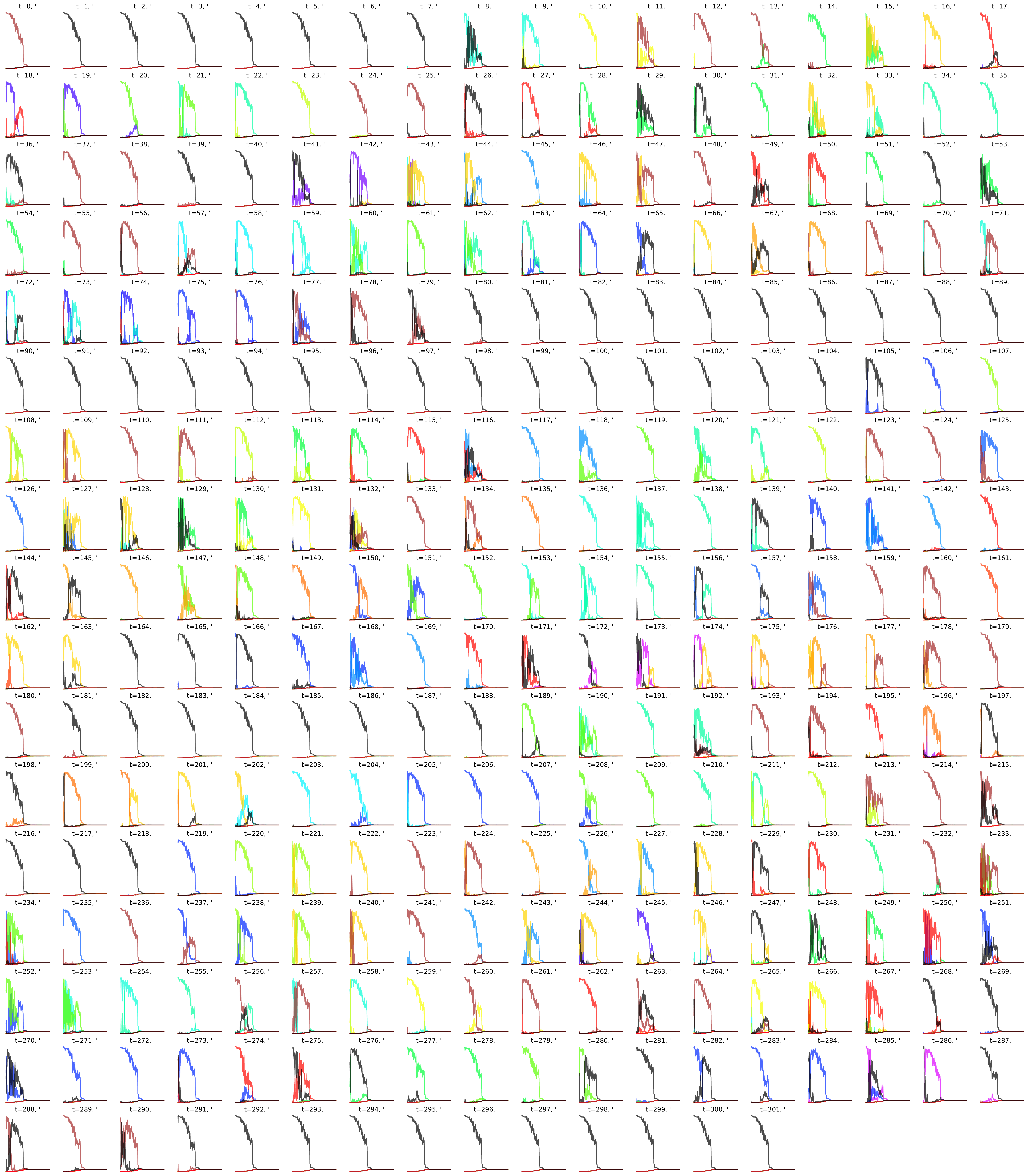}
    \includegraphics[width=\textwidth]{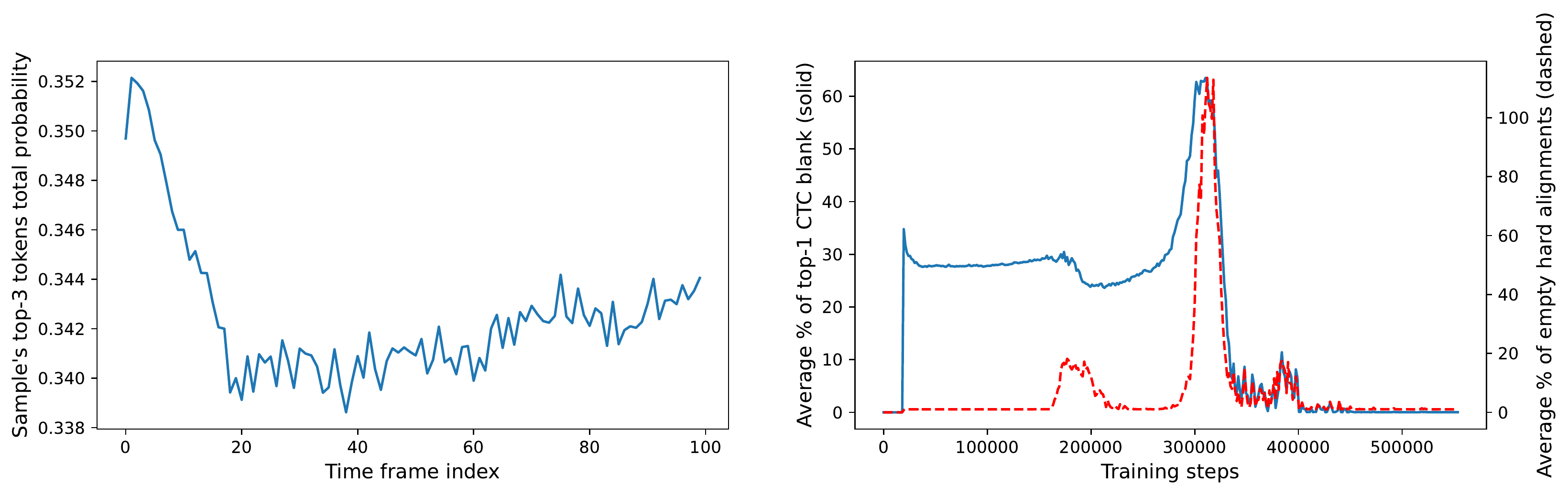}
    \caption{Training dynamics with soft-labels and target prior when divergence to non-blank distribution happens. We use 100h labeled data, $C=100, p=0.1, \tau=10, \beta=1, \gamma=0.05$.}
    \label{fig:per-frame-6}
\end{figure}

Below in Figures~\ref{fig:per-frame-1},\ref{fig:per-frame-2},\ref{fig:per-frame-3},\ref{fig:per-frame-4},\ref{fig:per-frame-5},\ref{fig:per-frame-6} we show detailed per-frame token distribution evolution during training with 100h labeled data, $C=100, p=0.1$. Top part of these Figures displays evolution in training of output token distribution per frame for one particular random sample from unlabeled set~$U$. Every sub-figure is a time frame $t$ of the sample. For every sub-figure: $x$-axis is a training step when we trained on the sample, $y$-axis is the probability of a token in range $[0, 1]$. CTC blank token is denoted by \#. 
Bottom left part shows total probability of top-3 tokens per frame for set of unlabeled samples averaged across training steps, while bottom right shows percentage of CTC blank token in hard-path PLs (solid) and percentage of empty hard-alignment PLs (dashed) during training.
\clearpage

\begin{figure}[ht!]
    \centering
    \includegraphics[width=\textwidth]{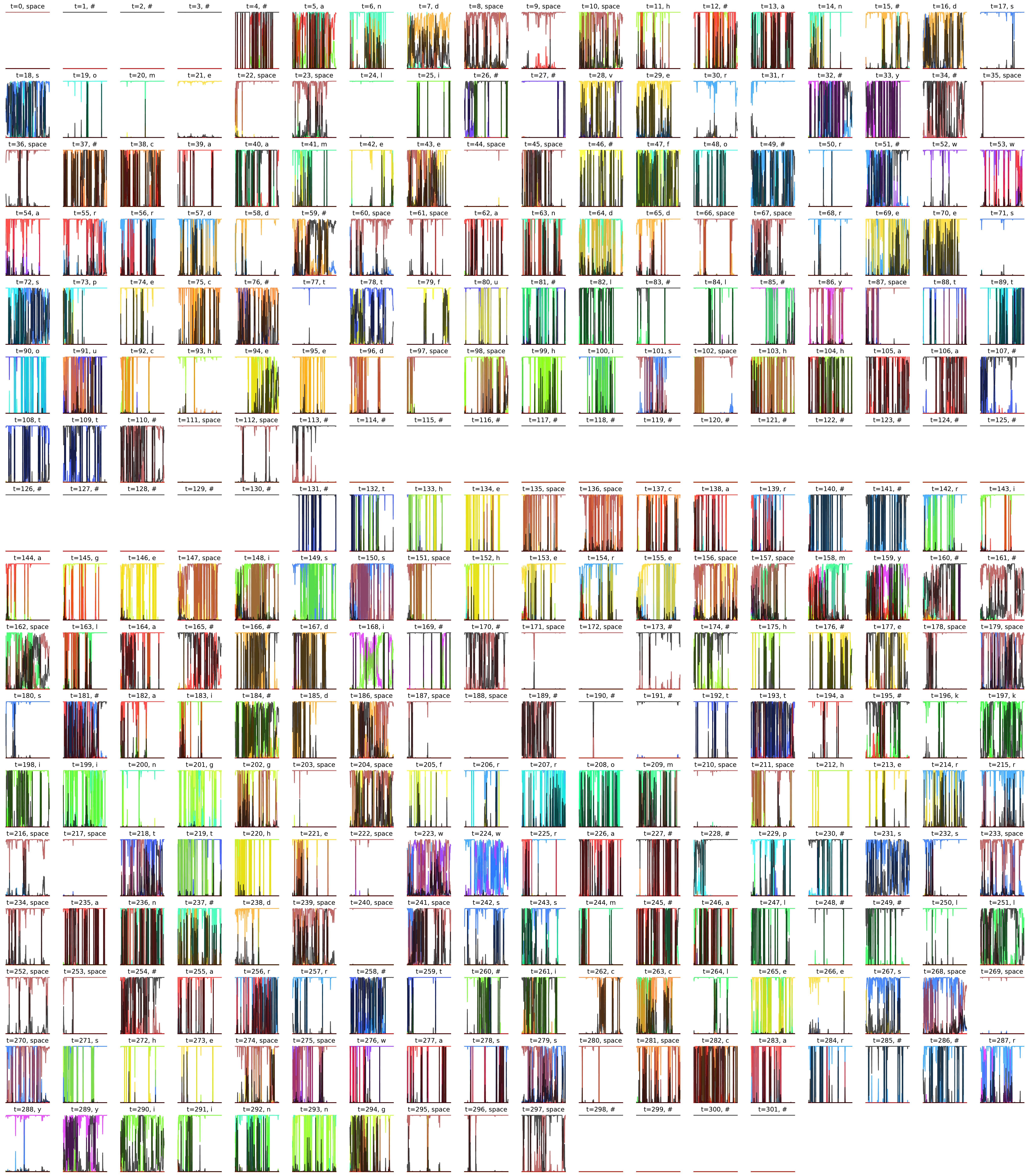}
    \includegraphics[width=\textwidth]{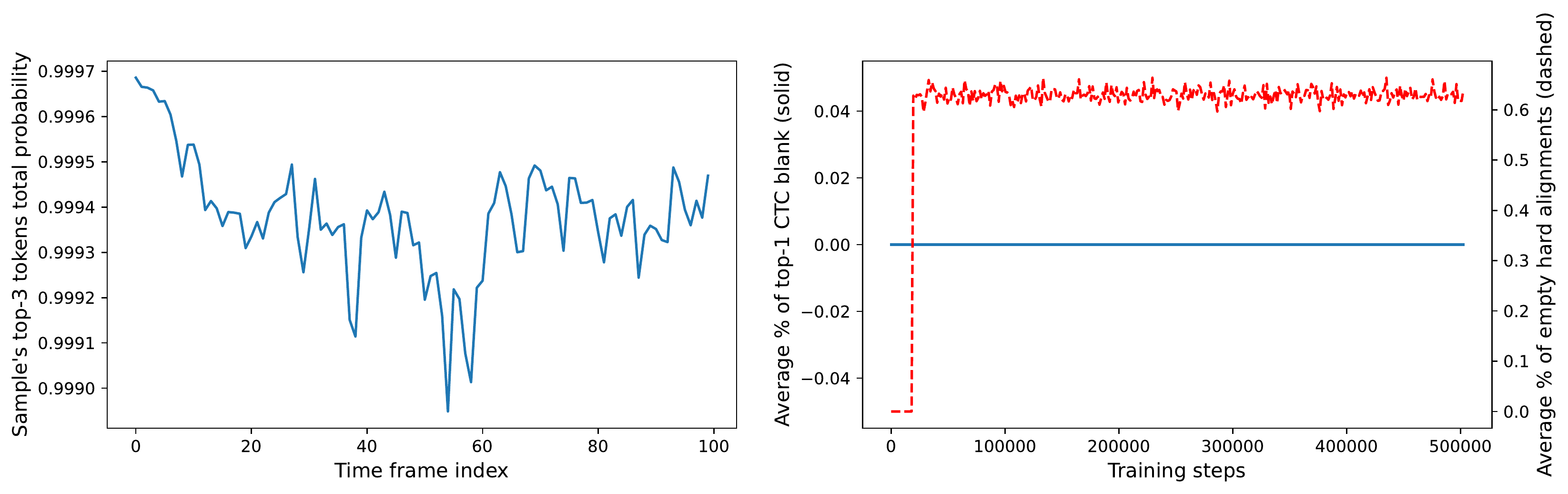}
    \caption{Training dynamics for hard-path with 100h labeled data, $C=100, p=0.1$.}
    \label{fig:per-frame-1}
\end{figure}

\clearpage

\begin{figure}[ht!]
    \centering
    \includegraphics[width=\textwidth]{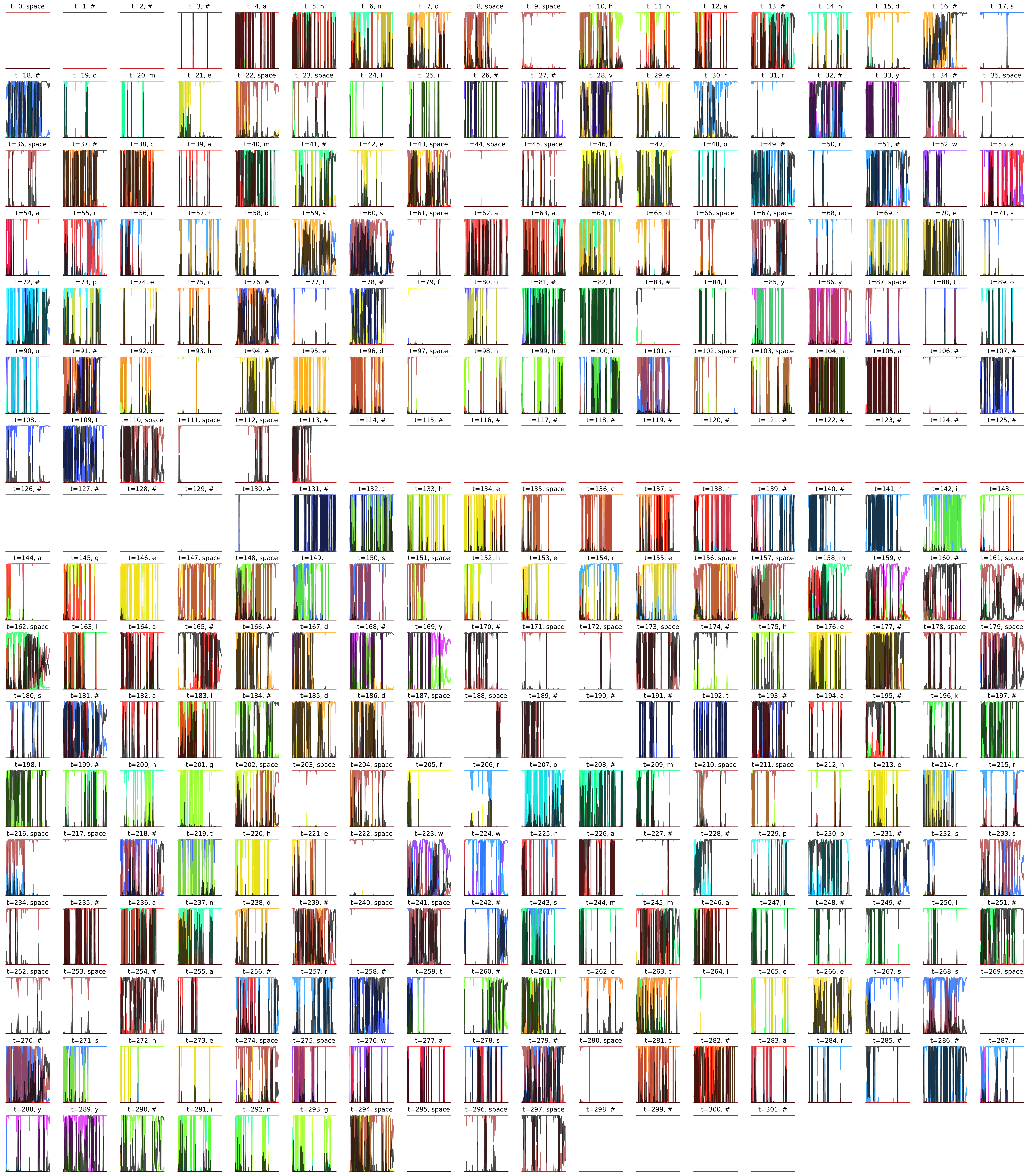}
    \includegraphics[width=\textwidth]{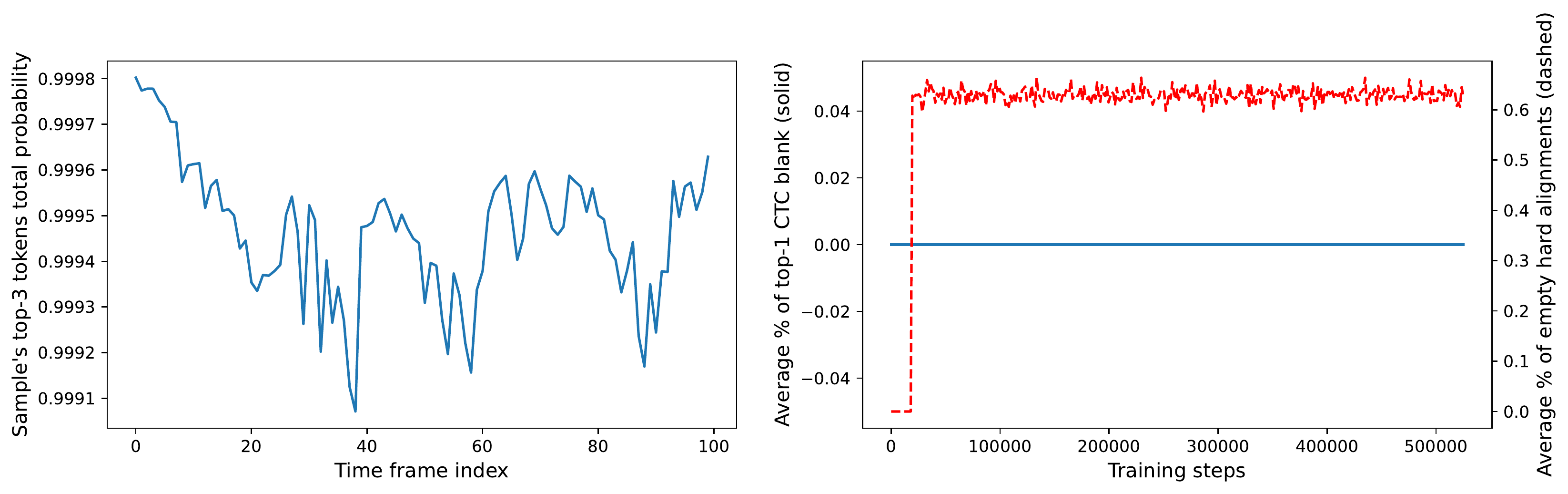}
    \caption{Training dynamics for sampling with 100h labeled data, $C=100, p=0.1, \tau=0.4$.}
    \label{fig:per-frame-2}
\end{figure}

\clearpage

\clearpage

\clearpage

\clearpage

\clearpage

\section{Collapsing to a Degenerate Solution and Labeled Set Memorization}\label{sec:app-overfitting}
\begin{figure}[h!]
    \centering
    \includegraphics[width=0.6\textwidth]{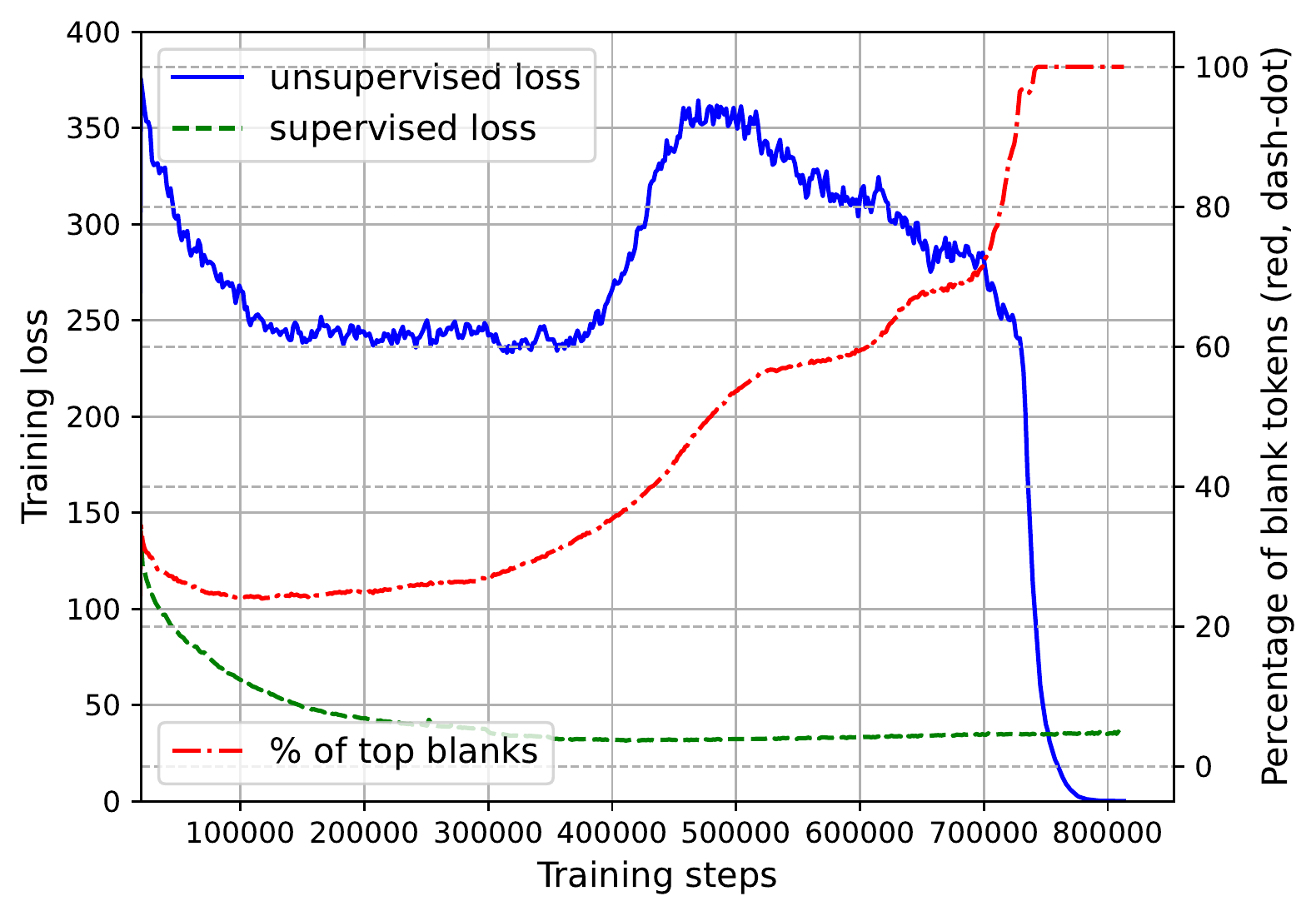}
    \caption{Example of a model trained with soft-labels that diverged during very long training to blank only token distribution (the same model as in Figure~\ref{fig:per-frame-5}). By the time of this convergence to a trivial solution the model has perfect zero loss on unlabeled data (solid, blue) while having non-zero and still decreasing training loss on labeled data (dashed, green) what indicates memorization of labeled data. The percentage of top blank tokens is presented too (dashed-dot, red).}
    \label{fig:memorization}
\end{figure}

For some models where the WER between hard-alignment and golden transcription increases, we keep training for a very long time. In all cases, while we observe different dynamics, all models converge to some degenerate solution, while memorizing the labeled training set. 
Convergence to a degenerate solution can be very slow and may take a huge number of training steps.
For all experiments we observe in some cases tendency to converge to blank prediction only (Figure~\ref{fig:per-frame-5}) or to e.g. uniform distribution (Figure~\ref{fig:per-frame-6}).
In case of convergence to these trivial solutions we observe that the training loss on pseudo-labeled data becomes zero while the loss on labeled data is non-zero (and decreasing), as shown in Figure~\ref{fig:memorization}. When we look at the model's generated transcriptions, we see empty transcriptions for unlabeled data and close to ground truth (WER is a few percentages) for labeled data. This indicates that the model is in the regime of labeled data memorization.

\begin{figure}[ht!]
    \centering
    \includegraphics[width=\textwidth]{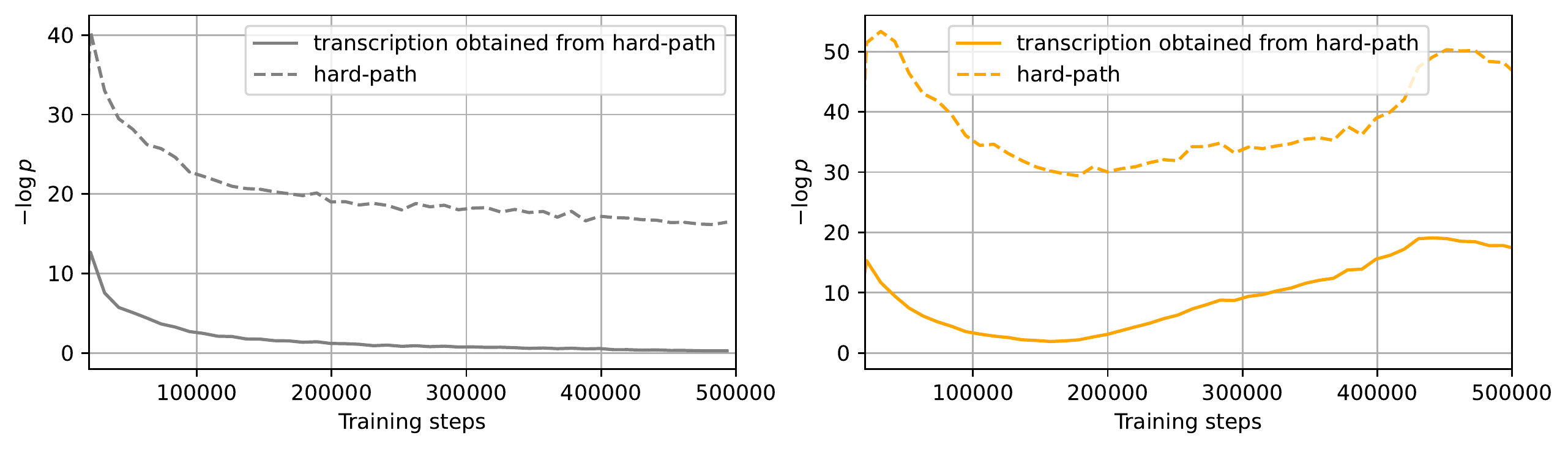}
    \caption{Comparison between hard-PL (left) and soft-PL (right) for 100h supervision, $C=100, p=0.1, N_U=3, \tau=10, \beta=1$. We generate hard-path pseudo-label for a sample and compute this hard-path probability and transcription (to which this hard-path leads) probability from the logits of the current model state. We can see that one-path itself never dominates and there are other alignments which lead to the same transcription and have significant probability mass.}
    \label{fig:ctc-loss-pl}
\end{figure}

\begin{figure}[ht!]
    \centering
    \includegraphics[width=0.7\textwidth]{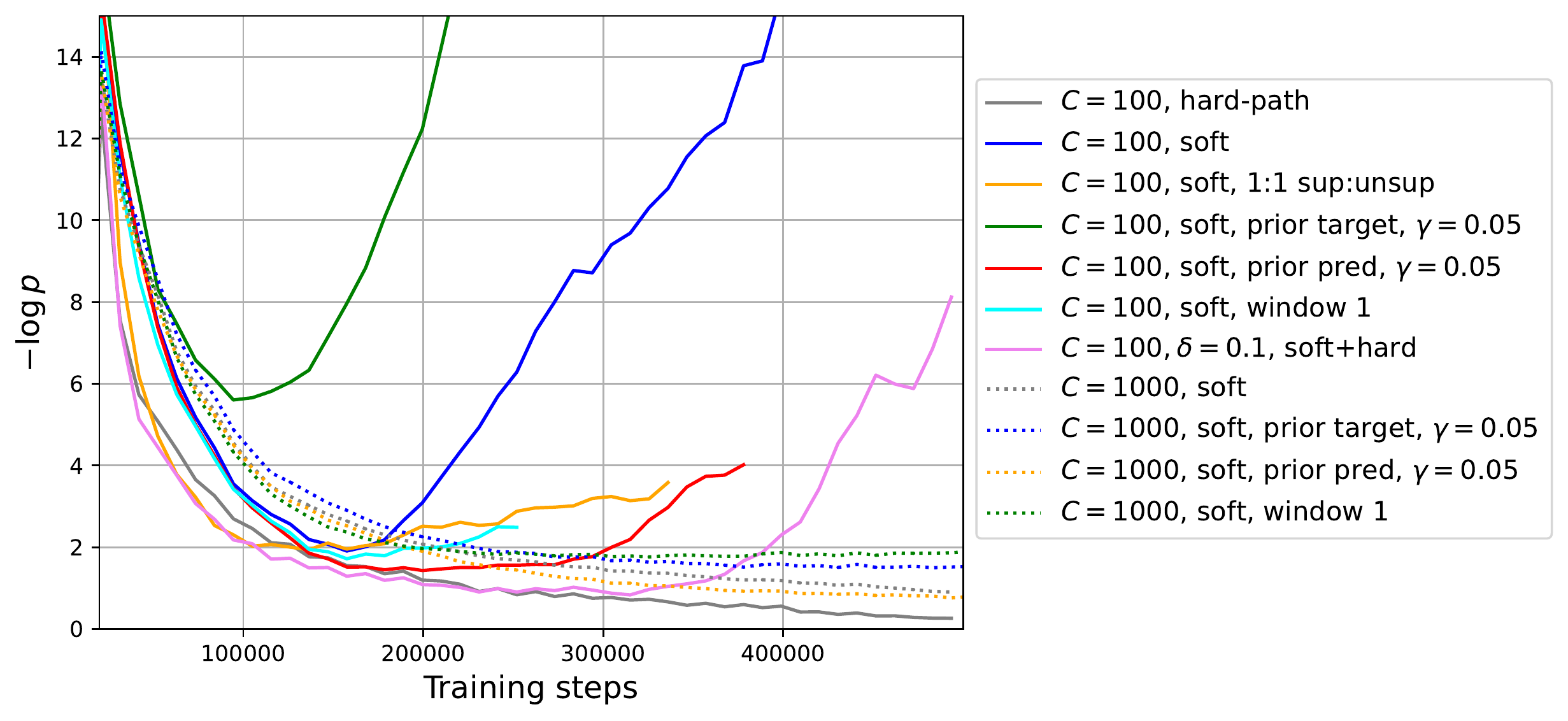} \label{fig:ctc-loss-pl-all}
    \caption{Comparison between different pseudo-labels for 100h supervision, $p=0.1, N_U=3, \tau=10, \beta=1$ for models from Figure~\ref{fig:instability}. We generate hard-path from a pseudo-label for a sample and compute this hard-path probability from the logits of the current model state. We can see that best one-path  probability decreasing ($-\log p$ is also growing) for unstable models.}
\end{figure}

\section{Ablations}\label{sec:app-ablations}
\subsection{Labeled Data Exclusion from PL Phase}
To confirm experimentally that there is no optimization problem (leading to trivial solutions) coming from the joint usage of different losses (CTC for labeled data, cross-entropy for unlabeled data), we stop to train on labeled data as soon as the cache is filled. With a cache size set to 100, we again observe model divergences with soft-labels, while cache size of 1000 is able to stabilize training. Both cases have worse performance than their hard-path counterpart with the same hyper-parameters.

\subsection{Teacher-Student Training as a Stable Soft-Labeling}\label{sec:app-teacher-student}
We tried also to disentangle continuous PL from soft-labels, and performed ablations with teacher-student PL and soft-labels. We first train a supervised only model on 100h of \librispeech{} (gets 18\% WER on \devother{}), use it to generate soft-labels for all unlabeled data and then train a new model from scratch on joint labeled and pseudo-labeled data. This model reduces WER on \devother{} to 14.1-14.3\% WER for different values of $\tau$ and $\beta$ while being very stable in training. We perform then the second round of teacher-student training using the latter model as a new teacher. Training a new model from scratch gives on \devother{} 12.8-13.4\% WER being stable again. Hence, we conclude that unstable training with soft-labels may come from continuous training dynamics. It is still an open question how to make continuous pseudo-labeling stable with soft-labels.

\subsection{Regularization: Blending Soft-Labels with Hard-Labels}\label{sec:app-blending}

\begin{figure}[ht!]
    \centering
    \includegraphics[width=\textwidth]{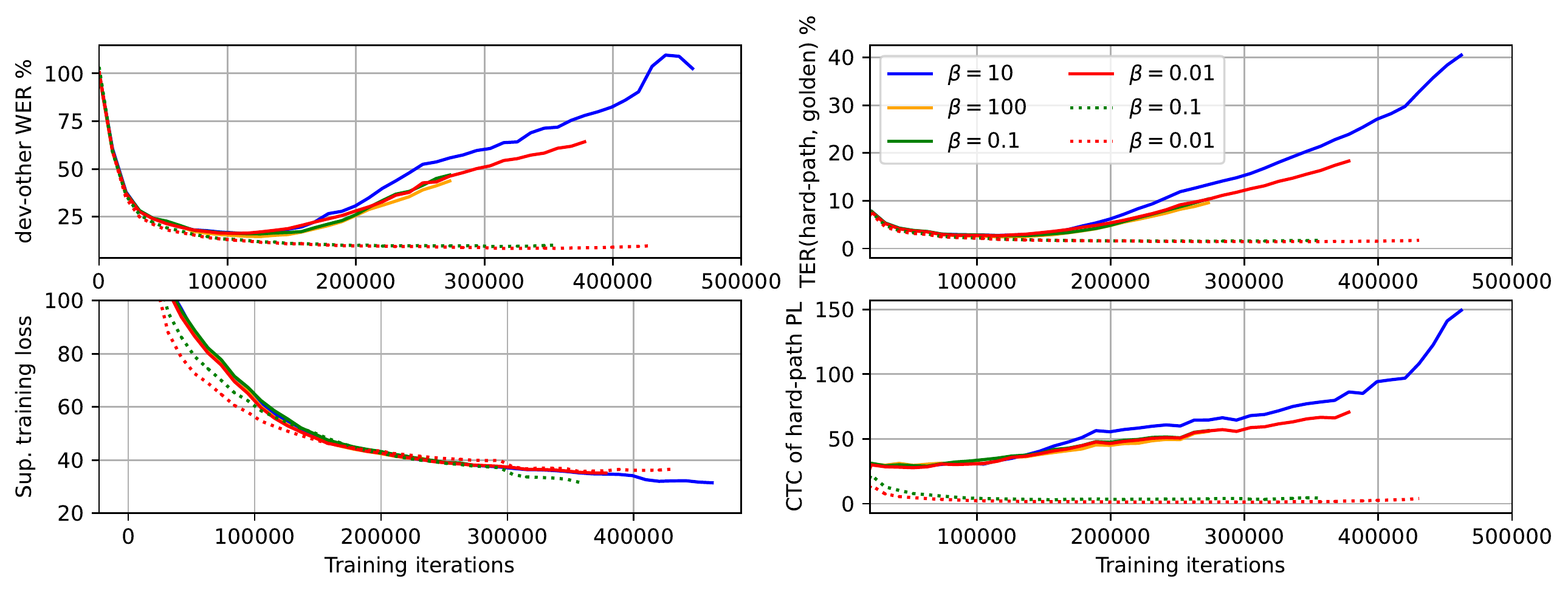}
    \caption{Comparison between soft-PL (solid) and blending soft-PL with hard-labels with $\delta=0.1$ (dashed). Training is done with temperature $\tau=1$ and 100h of labeled data, with cache parameters $C=100, p=0.1$.}
    \label{fig:blending-stability}
\end{figure}

\subsection{Sequence-Level Consistency: Pseudo-Labels Filtering}\label{sec:app-filetring}

\begin{figure}[ht!]
    \centering
    \includegraphics[width=\textwidth]{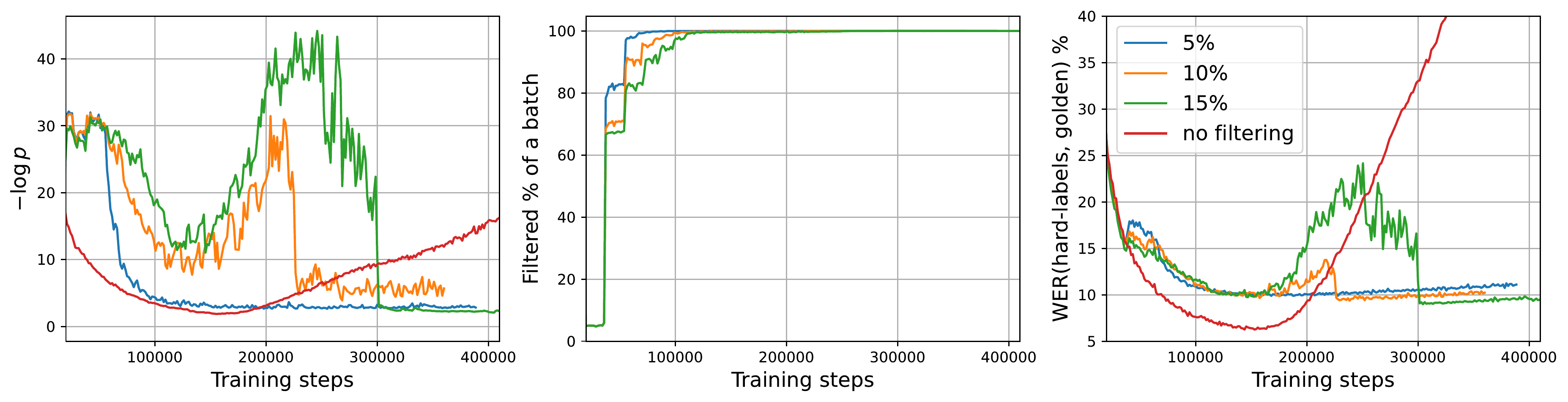}
    \caption{Training behaviour for PL filtering for sequence level consistency. We report three different thresholds on token distance between PLs across two consecutive training steps (5\%, 10\%, 15\%) and the baseline soft-labeling without any filtering (red) for $C=100, p=0.1, \tau=10, \beta=1$. Left: log probability of hard-path transcription of PLs, middle: percentage of samples filtered from the batch, right: WER between PL hard-path and golden transcription.}
    \label{fig:filtering-stability}
\end{figure}

As we discovered issues with sequence-level consistency for soft-labels (Figure~\ref{fig:pl-change-hard} vs Figure~\ref{fig:pl-change-soft}) for continuous training we tried regularization based on simple filtering pseudo-labeled data which breaks this consistency: we measure Levenshtein distance (on the token level) between hard-path transcriptions of current pseudo-label for a sample $\x\in U$ and {\it of previously used pseudo-label to make a training step} of the same sample; if the distance is larger than some threshold we filter a sample from the unlabeled batch for the current update step. Our preliminary experiments show that this filtering is able to stabilize the training with a proper threshold, see Figure~\ref{fig:filtering-stability}, however it reaches quite bad performance yet.

\end{document}